\documentclass[review,times, authoryear]{elsarticle}
\usepackage{graphicx} 
\usepackage[utf8]{inputenc}
\usepackage{amsmath}
\usepackage{amsfonts}
\usepackage{amssymb}
\usepackage{amsthm}
\usepackage{mathtools}
\usepackage{xfrac}
\usepackage{faktor}
\usepackage[english]{babel}
\usepackage{url}

\usepackage{float}

\usepackage{lineno}
\usepackage[most]{tcolorbox}
\tcbuselibrary{listings,breakable}
\tcbuselibrary{listingsutf8}

\usepackage{placeins}
\usepackage{graphicx}
\usepackage{subcaption}
\usepackage{algorithm}
\usepackage{subcaption}
\usepackage{algpseudocode}
\usepackage[capitalise]{cleveref}
\usepackage{booktabs}
\usepackage{geometry}
\journal{Smart Agricultural Technology}
\begin{document}

\begin{frontmatter}
\title{RowDetr: End-to-End Crop Row Detection Using Polynomials}

%

\author[1]{Rahul Harsha Cheppally\corref{cor1}}
\author[1]{Ajay Sharda}

\affiliation[1]{organization={ Department of Biology and Agricultural Engineering}, 
            addressline={Kansas State University}, 
            city={Manhattan}, 
            postcode={66502}, 
            state={KS}, 
            country={USA}}

\cortext[cor1]{Corresponding author}
\ead{asharda@ksu.edu}
\begin{abstract}
Crop row detection enables autonomous robots to navigate in gps denied environments. Vision based strategies often struggle in the environments due to gaps, curved crop rows and require post-processing steps. Furthermore, labeling crop rows in under the canopy environments accurately is very difficult due to occlusions. This study introduces RowDetr, an efficient end-to-end transformer-based neural network for crop row detection in precision agriculture. RowDetr leverages a lightweight backbone and a hybrid encoder to model straight, curved, or occluded crop rows with high precision. Central to the architecture is a novel polynomial representation that enables direct parameterization of crop rows, eliminating computationally expensive post-processing. Key innovations include a PolySampler module and multi-scale deformable attention, which work together with PolyOptLoss, an energy-based loss function designed to optimize geometric alignment between predicted and the annotated crop rows, while also enhancing robustness against labeling noise. RowDetr was evaluated against other state-of-the-art end-to-end crop row detection methods like AgroNav and RolColAttention on a diverse dataset of 6,962 high-resolution images, used for training, validation, and testing across multiple crop types with annotated crop rows. The system demonstrated superior performance, achieved an F1 score up to 0.74 and a lane position deviation as low as 0.405. Furthermore, RowDetr achieves a real-time inference latency of 6.7ms, which was optimized to 3.5ms with INT8 quantization on an NVIDIA Jetson Orin AGX. This work highlighted the critical efficiency of polynomial parameterization, making RowDetr particularly suitable for deployment on edge computing devices in agricultural robotics and autonomous farming equipment.
\end{abstract}

\end{frontmatter}

\section{Introduction}

The application of autonomous robots in high-throughput phenotyping has experienced a significant surge in recent years, driven by the need for precision and efficiency in agricultural tasks \cite{Atefi2021-dw}. These advanced robotic systems are transforming the field by automating the complex task of phenotyping with unparalleled precision. Although autonomous solutions in agriculture have been explored for decades \cite{9243253}, recent advances have pushed the limits of this technology, particularly in addressing challenges related to GPS-denied navigation in dense crop environments.

For nearly two decades, GPS-based autonomy has been the cornerstone of agricultural robotics. Studies such as \cite{kise2002enhancement} and \cite{gan2007implement} have demonstratedted the use of RTK and GPS-based systems to guide tractors and harvesters with high precision. The GPS coordinates of rows can either be determined during planting or estimated using systems such as \cite{yan_seed_mapping, CHEPPALLY2023100182}, which provide accurate crop and row location data. However, these GPS-based technologies face critical limitations under dense canopy cover, where GPS multipath errors or complete signal loss render them unreliable. For example, the RoboBotanist project highlighted these challenges, encountering significant navigation difficulties beneath dense canopies due to GPS interference from leaf coverage \cite{7989418}. These challenges emphasize the need for alternative solutions designed specifically for under-canopy navigation, where GPS-based methods fall short.

Vision-based approaches have emerged as a promising alternative, combining object detection, visual odometry, and sensor fusion \cite{ctx23878273330002401, Sivakumar_2021}. However, as noted in \cite{sivakumar2024lessons}, these methods encountered challenges in scenarios involving discontinuous crop rows or curved row structures.
This highlights the critical need for more accurate and robust methods for detecting crop rows.

Traditional methods for crop row detection relied on hand-crafted features, ROI or Hough transforms to detect straight rows \cite{7003860, JIANG2016211, Yuan, tasseled}. Similar to traditional methods of object detection \cite{dalal2005histograms} prior to deep learning they would need to be tunned for each application. Recent advancements in learning based crop row detection methods include the development of an end-to-end detection network by \cite{LI2023120345}, which predicts line segments for rows. While effective for straight rows, this method encounters difficulties with curved rows due to its reliance on linear representations. Similarly, \cite{yang2023real} employed YOLO and bounding box extraction for boundary detection, achieving a latency of 25 ms but requiring additional post-processing steps. Segmentation-based methods, such as those proposed by \cite{YU2023107811, CRDLD}, rely on full-image segmentation for row detection, achieving frame rates of 15–17 fps. However, these methods face significant challenges in real-time applications due to computational inefficiencies in post-processing, which also diminish accuracy in dense foliage environments. While these approaches address certain aspects of row detection, they remain limited in their ability to handle curved rows and dense canopy conditions.

A parallel study by \cite{RowColRowDetection} seeks to address the aforementioned limitations by representing crop rows as a series of points and employing an end-to-end method for row detection and eliminates the need for post processing.
However, as highlighted in their findings, this approach also faces challenges in managing discontinuities in crop rows. While non-learning approaches, including UAV-based systems \cite{9255701, 9885256}, have been explored for curved row detection, end-to-end learning-based methods tailored for under-canopy conditions are still largely unexplored. This gap motivates the development of robust frameworks capable of handling label noise, canopy density, and curved structures, while remaining computationally efficient for edge deployment.

In contrast, the field of autonomous driving has extensively explored lane detection, with particular emphasis on polynomial regression methods for real-time performance. 
Techniques proposed by \cite{yang2024polylanenet++, tabelini2020polylanenet, 9624545} model lanes as polynomial functions, achieving lower latency compared to segmentation-based approaches.
However, this method is not eaisly transferable to agricultural row detection. Due to lack of attaining good annotations for dense canopies.
Crop canopies are dense, not very structured and clear in the images compared to the lane lines.
Regression loss is not very effective for handling noisy labels \cite{liu2022robustobjectdetectioninaccurate, barron2019generaladaptiverobustloss}.
Therefore, the objective of this study was designed with the following objectives:
\begin{enumerate}
\item Develop a Robust Loss Function to handle noisy labels and label ambiguities.
\item Design an end-to-end framework for row detection in dense crop canopies, optimized for computational efficiency on edge devices.
\item Collect a real-world dataset to evaluate the model's performance and compare it against existing state-of-the-art end-to-end methods.
\end{enumerate}

\begin{figure}[!ht]
    \centering
    \includegraphics[width=\textwidth]{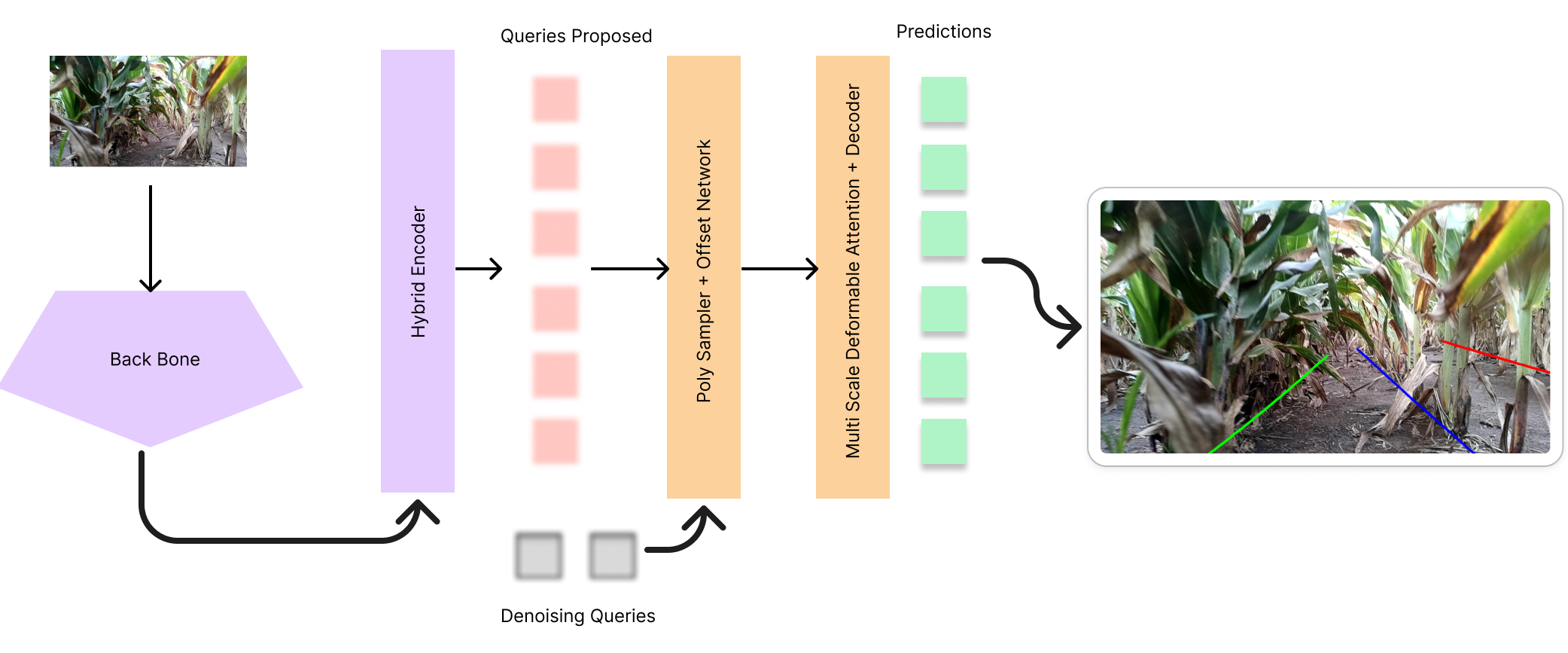}
    \caption{Overview of the proposed architecture. The input image is first processed by a backbone to extract multi-scale features, which are then passed through a hybrid encoder. The encoder proposes a set of top queries, while denoising queries are also introduced to stabilize training. A polynomial sampler, followed by an offset network, refines the proposed queries, enabling structured sampling along polynomial paths. These refined queries are passed to a multi-scale deformable attention decoder, which produces final predictions. The image on the right visualizes the predicted curves (red, green, and blue), each corresponding to a distinct row hypothesis.}
    \label{fig:model-arch-1}
\end{figure}

\section{Methodology}

\subsection{Dataset}
\label{sec:dataset}

The dataset employed to train the proposed Row Detection Framework (Fig.~\ref{fig:model-arch-1}) comprises multiple subsets acquired under diverse conditions using various devices, thereby ensuring broad scenario coverage. Table. \ref{tab:dataset-devices} illustrates the distribution of images across these subsets.
\begin{table}[ht]
    \centering
    \caption{Dataset Summary with Devices Used for Collection. V4, V5 and V6 represent vegetative stages of growth of corn.}
    \label{tab:dataset-devices}
    \resizebox{\textwidth}{!}{

    \begin{tabular}{lccll}
        \hline
        \textbf{Dataset Name} & \textbf{Total Images} & \textbf{Device Used} & \textbf{Stage of Growth} & Resolution \\
        \hline
        Corn V5 & 1250 & Handheld devices (Samsung, iPhone) & Vegetative Stage (V5) & 1080x1920\\
        Corn V4 & 835 & Basler camera on metering stick & Vegetative Stage (V4) & 1200x1920\\
        Corn V6 & 795 & Basler camera on metering stick & Vegetative Stage (V6) & 1200x1920\\
        CRDL Dataset & 1930 & RealSense D435i & Refer to the source paper \cite{CRDLD} & 512x512 \\
        Sorghum Dataset & 2152 & Robot & Mature Stage \& Ready for harvest & 720x1280\\
        \hline
    \end{tabular}
    }

\end{table}
\subsubsection{Data Collection Setup}
Multiple devices were used to create a comprehensive dataset summarized in Table \ref{tab:dataset-devices}. The use of multiple devices was crucial to encourage generalization and robustness of the model.

\begin{figure}[ht]
    \centering
    \begin{subfigure}[t]{0.6\textwidth}
        \centering
        \includegraphics[width=\textwidth]{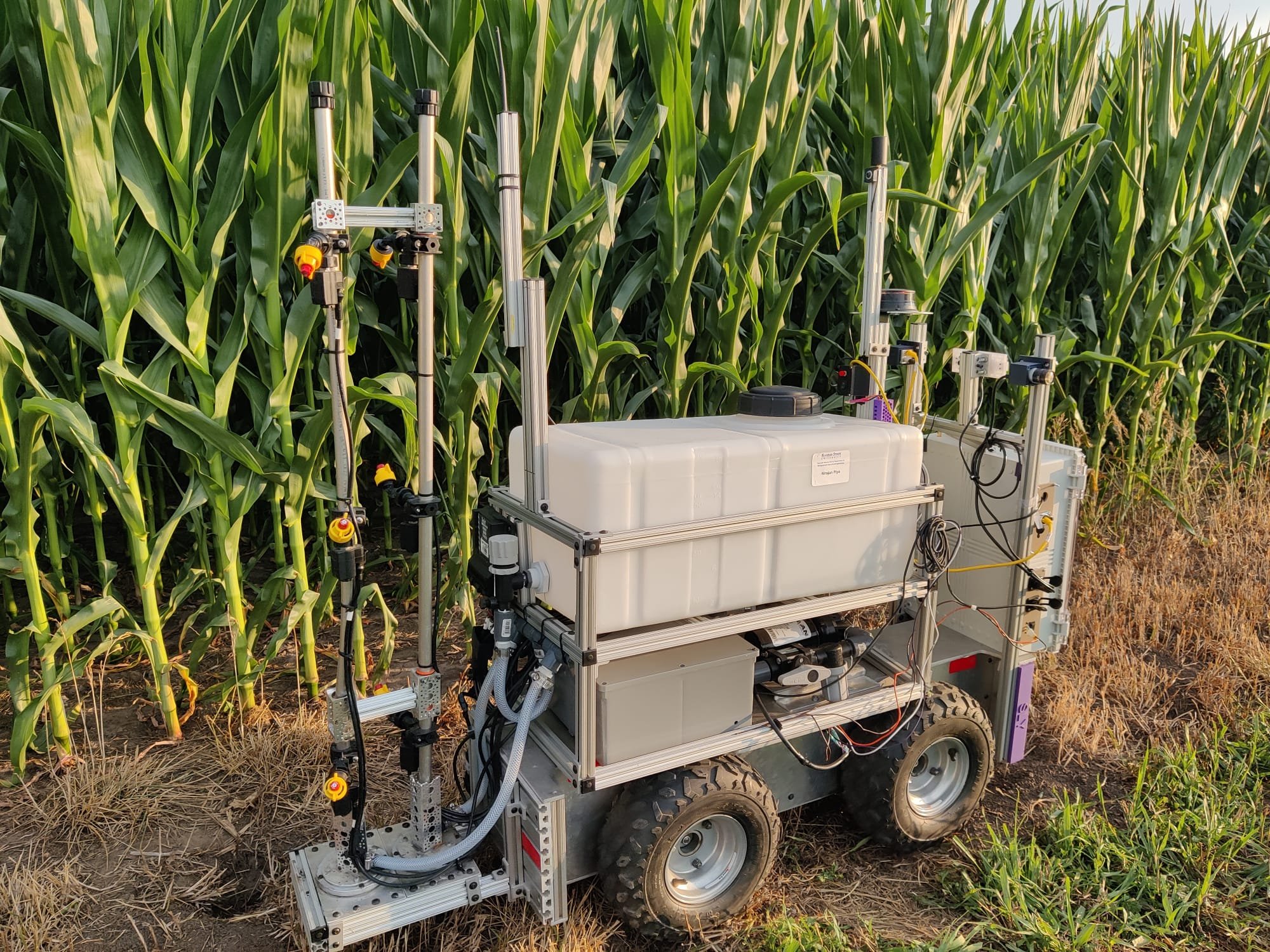}
        \caption{The under-the-canopy robot equipped with Jetson Orin and Luxonis OAK-D Pro for image acquisition.}
        \label{fig:sly}
    \end{subfigure}
    \hfill
    \begin{subfigure}[t]{0.3\textwidth}
        \centering
        \includegraphics[width=\textwidth]{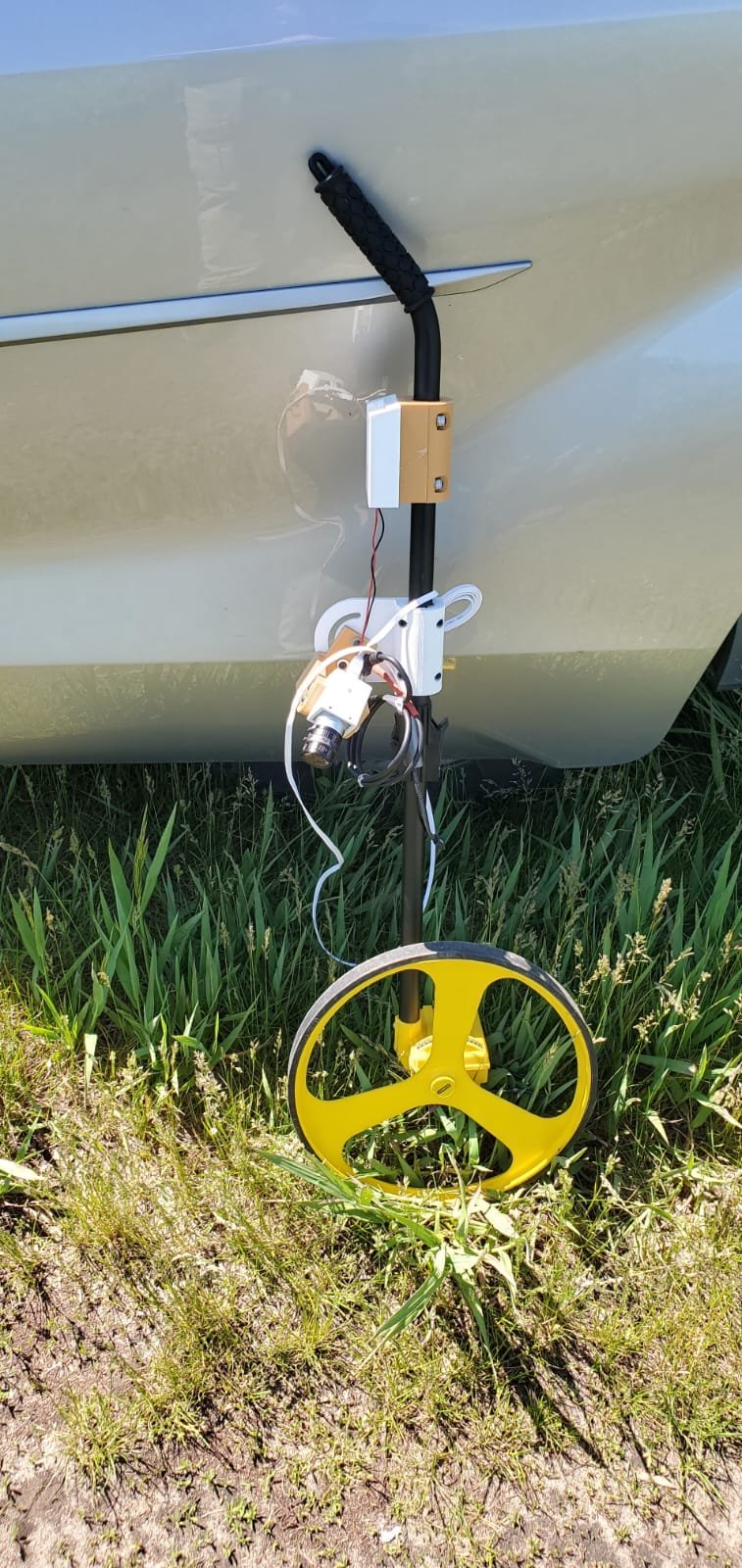}
        \caption{Metering Stick for height reference in dataset collection.}
        \label{fig:pogo-stick}
    \end{subfigure}
    \caption{Hardware setup used for data collection. The under-the-canopy robot (left) captures high-resolution images for row detection, while the metering stick (right) provides accurate height measurements.}
    \label{fig:robot_setup}
\end{figure}

\begin{itemize}
    \item \textbf{Custom under-the-canopy robot}: As shown in Fig. \ref{fig:sly}, this robot was equipped with a Jetson Orin \cite{jetson} and a Luxonis OAK-D Pro \cite{Luxonis} for image acquisition. RGB Images were collected using ROS2 and stored in Rosbags \cite{ROS2}. 
    \item \textbf{Metering stick}: As shown in Fig. \ref{fig:pogo-stick}, this device featured a Basler AC 1920-40mm Camera (Basler, Ahrensburg, Germany, Europe, 2024), which used pylon software on a Windows laptop.
    \item \textbf{Handheld devices}: Various handheld devices, including Samsung and iPhones, were also used for data collection.
\end{itemize}

\subsubsection{Dataset Composition}

\begin{figure}[!ht]
    \centering

    \begin{subfigure}[b]{0.4\textwidth}
        \includegraphics[width=\linewidth]{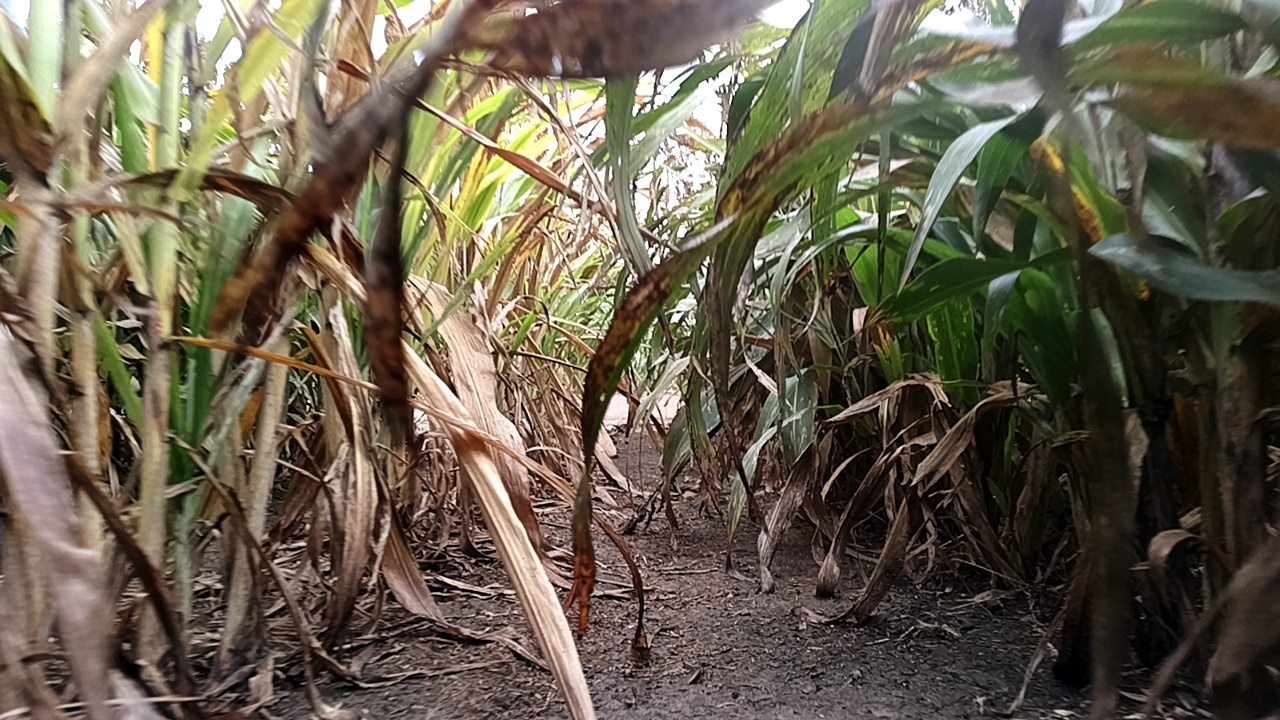}
        \caption{Sorghum canopy with curved rows collected using OAK-D Pro.}
        \label{fig:sorghum}
    \end{subfigure}
    \hfill
    \begin{subfigure}[b]{0.4\textwidth}
        \includegraphics[width=\linewidth]{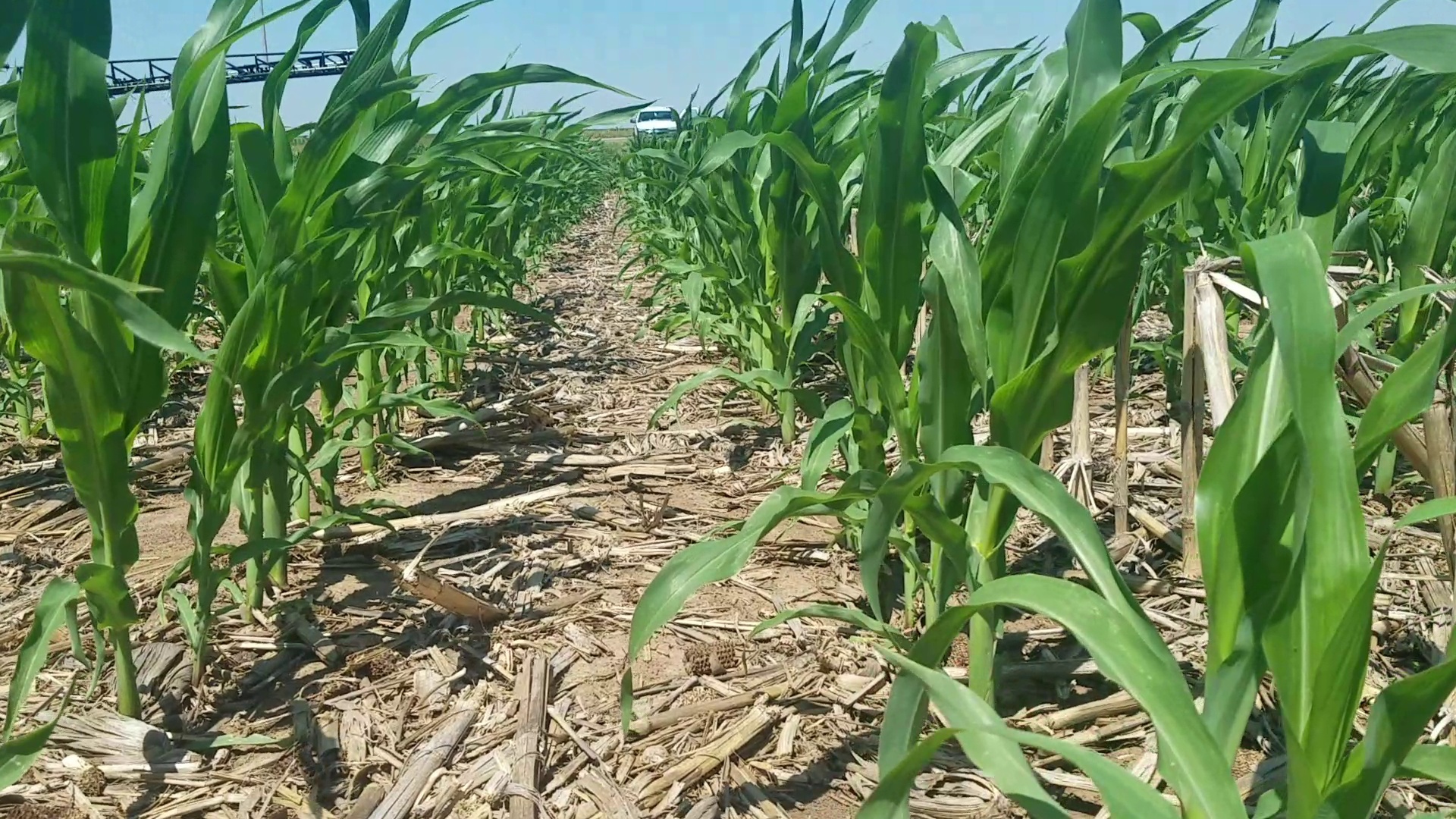}
        \caption{Corn V5 with surface residue (sample 1) Collected using Samsung, iPhone.}
        \label{fig:corn_v5}
    \end{subfigure}

    \vspace{1em}

    \begin{subfigure}[b]{0.4\textwidth}
        \centering
        \includegraphics[width=\linewidth]{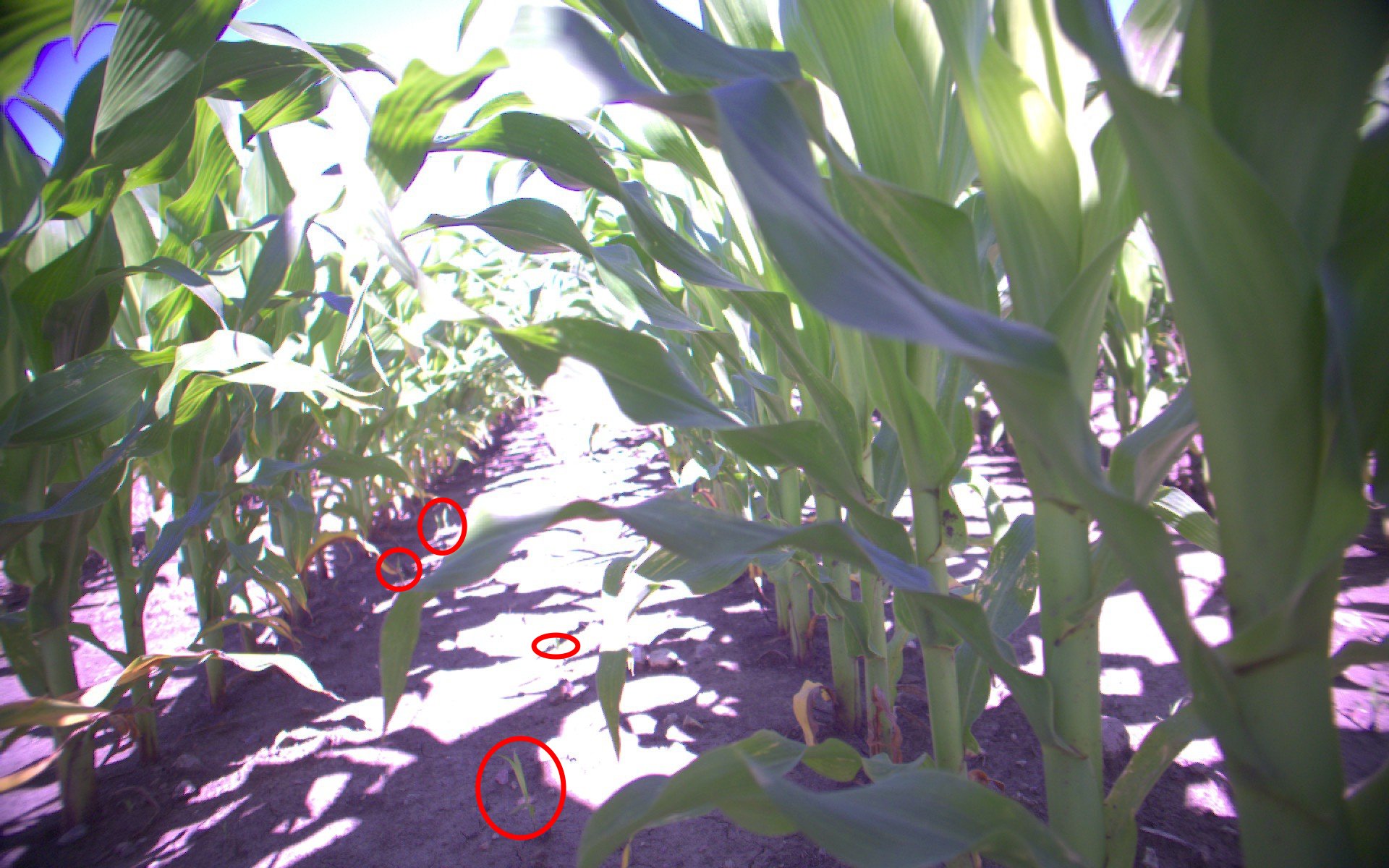}
        \caption{Corn V6 with visible weed presence annotated with a red ellipse. Collected using Basler camera on metering stick.}
        \label{fig:corn_v6}
    \end{subfigure}
    \hfill
    \begin{subfigure}[b]{0.4\textwidth}
        \includegraphics[width=\linewidth]{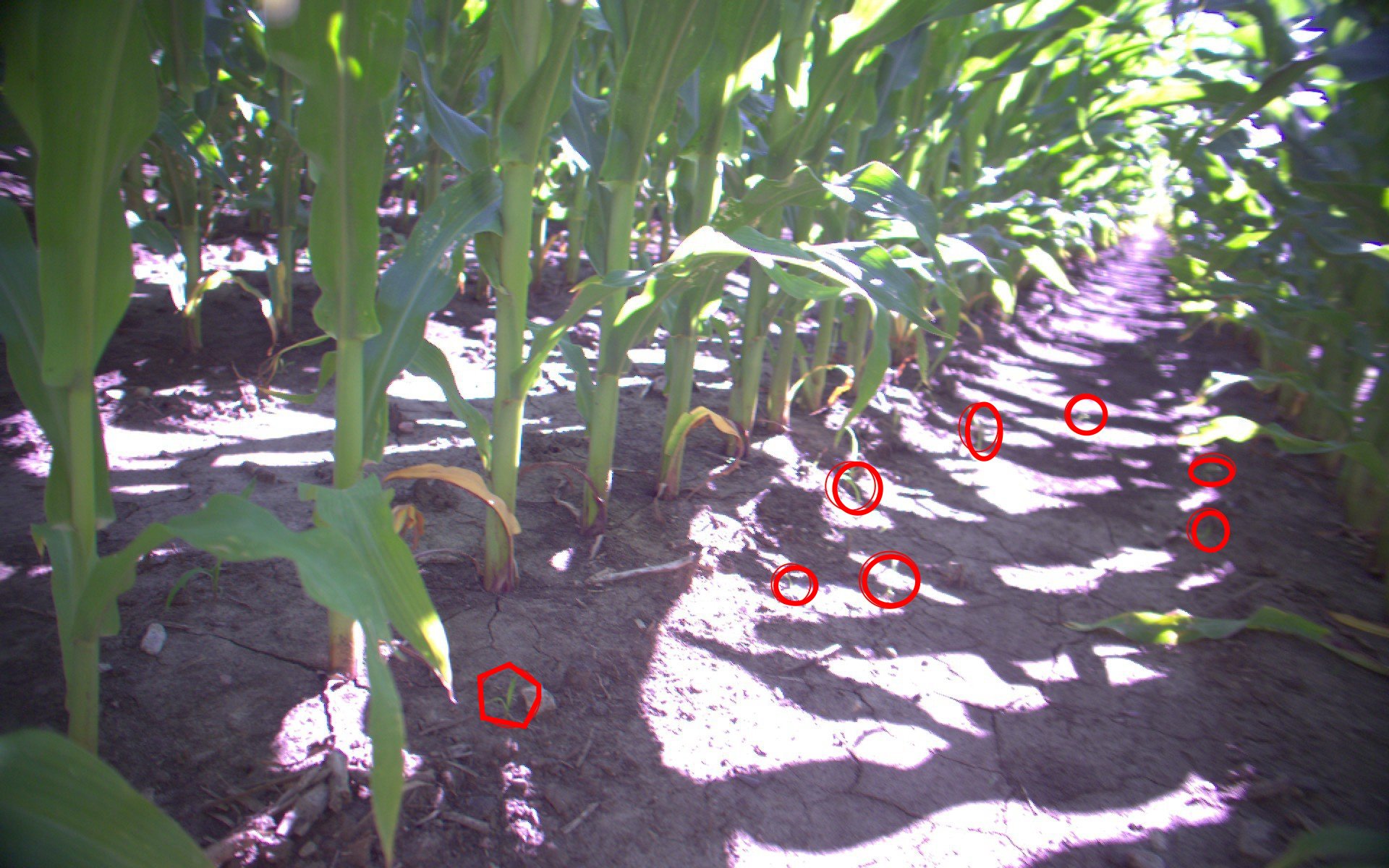}
        \caption{Corn V4 with weed growth weed annotated with a red circles. Collected using Basler camera on metering stick.}
        \label{fig:corn_v4}
    \end{subfigure}

    \vspace{1em}

    \begin{subfigure}[b]{0.4\textwidth}
        \includegraphics[width=\linewidth]{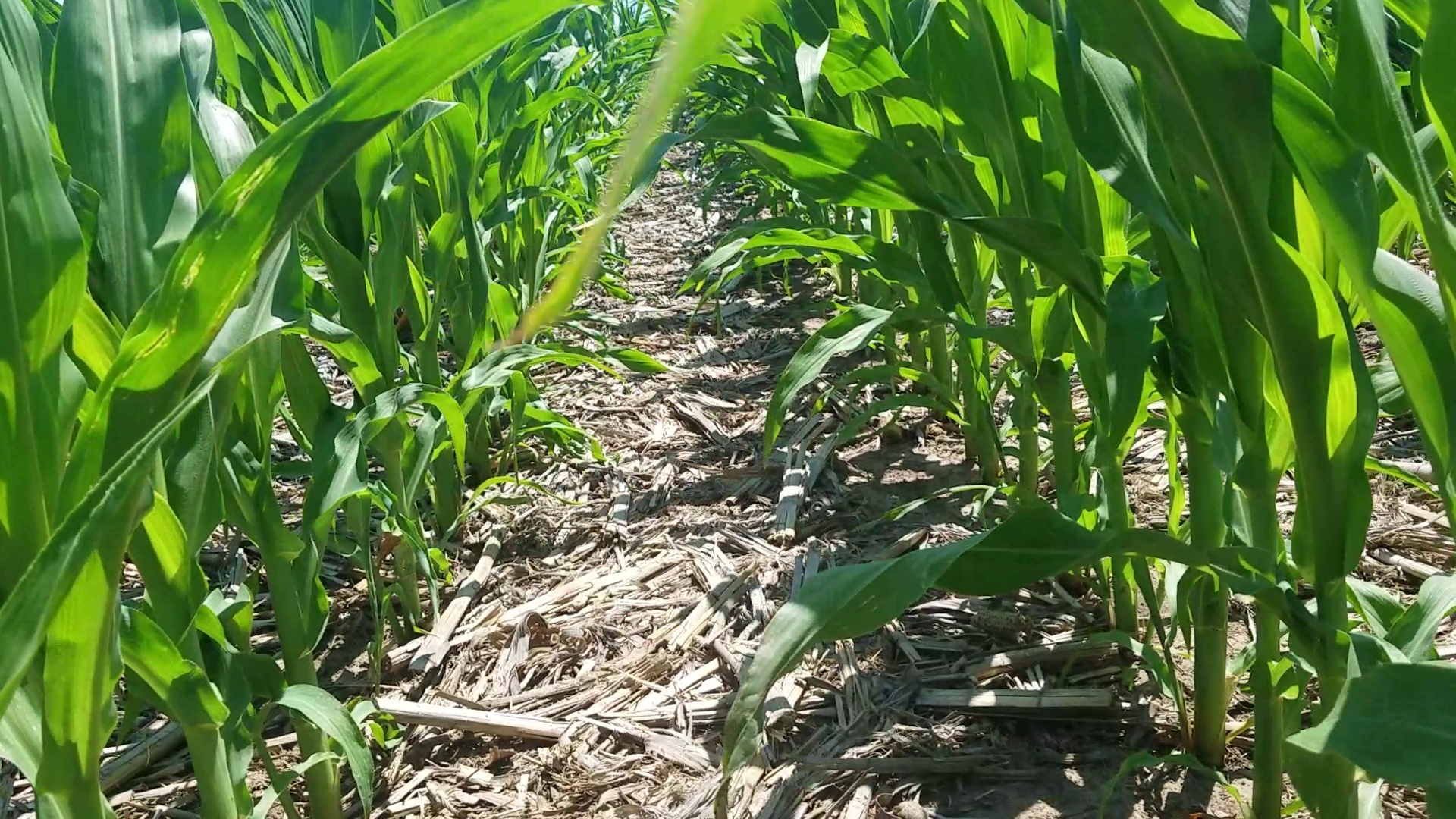}
        \caption{Corn V5 with surface residue (sample 2). Collected using Samsung, iPhone.}
        \label{fig:corn_v52}
    \end{subfigure}
    \hfill
    \begin{subfigure}[b]{0.3\textwidth}
        \includegraphics[width=\linewidth]{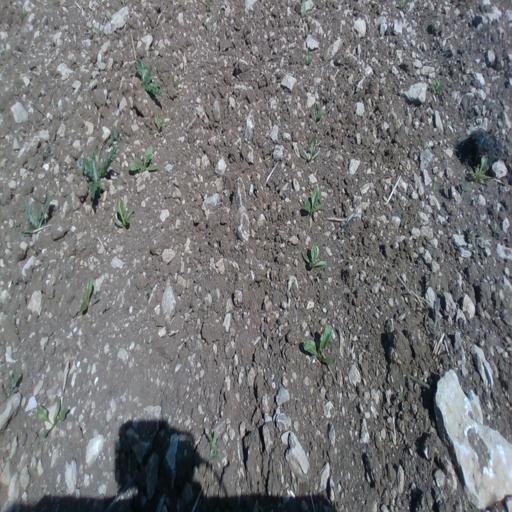}
        \caption{CRDL-Dataset Sample. Collected using RealSense.}
        \label{fig:crdl_img}
    \end{subfigure}

    \caption{Representative samples from the dataset illustrating various challenging field conditions: canopy occlusion (a), row curvature (a), weed presence (c), and surface residue (b, and d); including presence of weeds within the rows (e).}
    \label{fig:dataset_samples}
\end{figure}

The collected data was divided into several distinct subsets, each characterized by specific conditions. Table \ref{tab:dataset-devices} provides a summary of the image distribution across these subsets. All subsets used in this study were gathered under sunny or partly cloudy weather.None of the datasets were collected at night.
\begin{itemize}
\item \textbf{Corn V5} (1250 images): This subset captured over 45 minutes using handheld devices (Samsung, iPhone), with frames sampled every 2.16 seconds to minimize redundancy. The corn was in the V5 growth stage, with significant ground residue. Fig. \ref{fig:corn_v5} and \ref{fig:corn_v52} shows a sample from the Corn V5 subset.
\item \textbf{Corn V4 and V6} (835 and 795 images, respectively): These subsets were collected using the metering stick (Fig. \ref{fig:pogo-stick}). The camera recorded at 3 FPS with induced motion disturbances. The captured TIFF images were later converted to JPEG. Sample images from the Corn V4 and V6 subsets are shown in Fig. \ref{fig:corn_v4} and \ref{fig:corn_v6} respectively.
\item \textbf{CropRow Detection Lincoln Dataset (CRDLD)} (1250 training, 430 testing, 250 validation images): This dataset was adapted from the study \cite{CRDLD}, with only the RGB data being used for row detection. Fig. \ref{fig:crdl_img} shows a sample from the CRDLD subset.
\item \textbf{Sorghum Dataset} (2152 images): This subset collected using custom-under-canopy robot. RGB Images were extracted from the ros bag recordings of the robot navigating under a mature sorghum canopy, with every 25th frame (~0.8s interval) sampled to reduce redundancy. Fig. \ref{fig:sorghum} shows a sample from the sorghum subset.
\end{itemize}

\subsubsection{Test-Train Split}
\begin{figure}[ht]
        
    \centering
    \includegraphics[width=0.8\textwidth]{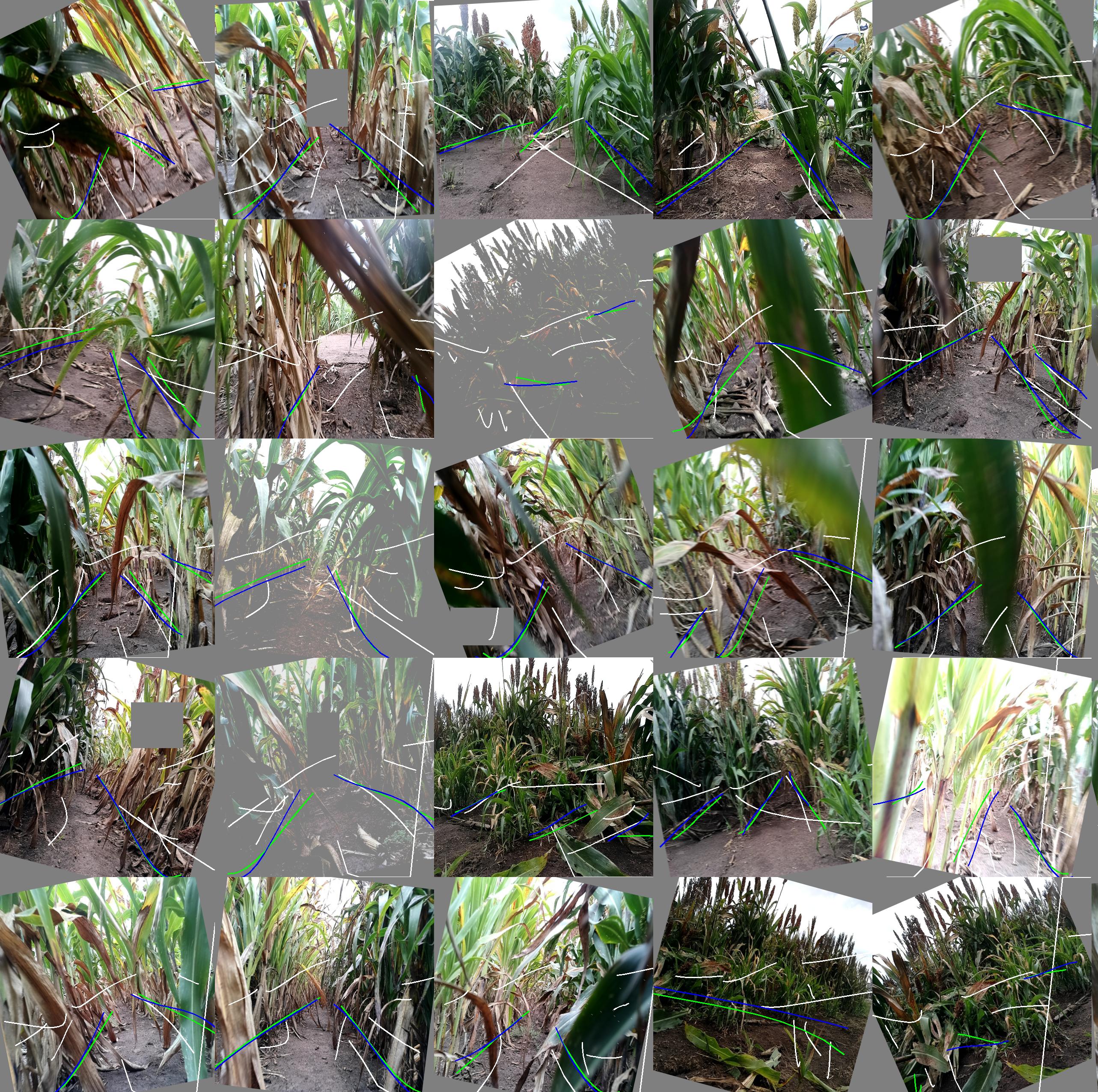}
    \caption{Samples from the training batch (Augmented): labeled samples (Blue), positive predictions with confidence greater than 0.5 (Green), and non-existent row predictions with confidence less than 0.5 (White).}
    \label{fig:detections}
\end{figure}

Samples from the dataset, shown in Fig. \ref{fig:dataset_samples}, highlight the primary challenges of real world agricultural environments. These include occlusions from dense canopy cover, curved row structures, weed interference, and surface residue, all of which complicate accurate row detection. The CRDL dataset retained its original training, validation, and test splits as defined in the source study \cite{CRDLD}, while the remaining datasets were randomly partitioned into 64\% training, 16\% validation, and 20\% testing subsets. To improve model robustness \cite{rebuffi2021dataaugmentationimproverobustness}, images were resized to 640×384, and augmented at training time using a variety of techniques implemented in the \texttt{kornia} library \cite{eriba2019kornia}. These augmentations included affine transformations, simulated weather effects, color jittering, motion blur, and random cutout. Fig. \ref{fig:detections} presents examples of augmented samples generated through this process.

\subsubsection{Annotation} \label{sec:labelling}

The annotation process involved marking plant stalks using the polyline tool in \cite{labelbox}. When stalks were not visible, labelers marked the boundary between the plant and the navigable path. Each crop row was annotated as a polyline with at least four points, enabling the fitting of second-degree polynomials. The annotated rows are shown as blue curves in Fig. \ref{fig:detections}. When multiple rows were visible, labelers were instructed to annotate all of them, so a single image could contain several labeled rows.

For the CRDLD dataset, a boundary-based labeling strategy was adopted since stalks were not consistently visible.
\subsection{End-to-End Architecture: RowDetr } \label{sec:RowDetr}

For the task of crop row detection, an architecture inspired by RT-DETR \cite{rtdetr} was implemented. This model, named RowDetr, utilizes a transformer based encoder decoder framework. While RT-DETR is designed for bounding box detection, RowDetr is uniquely tailored to detect and represent crop rows as polynomial curves.

As illustrated in Fig. \ref{fig:model-arch-1}, the RowDetr architecture consists of several key components: a backbone, a hybrid encoder, a polynomial proposal generation module, a PolySampler, an offset network, and multi-scale deformable attention. The model takes an image as input and outputs M polynomial proposals. Each proposal provides both a probability score and the specific parameters needed to represent a row, which are discussed in detail in Section \ref{sec:row_representation}.

\subsubsection{Row Representation}\label{sec:row_representation}

Lane detection studies, such as \cite{yang2024polylanenet++, tabelini2020polylanenet}, model predicted curves as polynomials of the form \( v = f(u) \), where \( f \) is a third-degree polynomial. This representation is effective for typical lane detection scenarios but becomes problematic for agricultural row detection, particularly when rows are perpendicular to the camera's field of view. This issue arises with top-down imagery from UAVs or other downward-facing camera setups, as well as with augmented data where rows are rotated.

To overcome this limitation, we model both the \(u\) and \(v\) coordinates as functions of a new parameter, \(\lambda\) (Eq. \ref{eq:lambda}). The row is first annotated with a series of \(n\) points, \(p_i = (u_i, v_i)\). To ensure a consistent monotonic relationship for the polynomial functions, these points are sorted by their \(v\)-coordinate before further processing.

The parameter \(\lambda\) is defined as the normalized cumulative chord length:
\begin{equation}
\label{eq:lambda}
\lambda_i = \frac{\sum_{j=1}^{i} \| p_{j-1} p_j \|}{\sum_{j=1}^{n} \| p_{j-1} p_j \|},
\end{equation}
where \(\lambda \in [0, 1]\). This parameterization allows us to represent the row with two separate polynomials of degree \(k\): \(u = U(\lambda)\) and \(v = V(\lambda)\).

\subsubsection{Backbone and Hybrid Encoder:}
Input images are first processed by a backbone network (A lightweight CNN like resnet18, efficientnet, regnetx\_008, resnet50 is used to enable real-time processing) to extract high-level features. These features are then passed to a hybrid encoder, where self-attention is applied exclusively on the final layer. A top-down Feature Pyramid Network (FPN) is incorporated into the encoder to integrate multi-scale features, ensuring the detection of rows at varying resolutions and distances.

\subsubsection{Polynomial Proposals and Query Selection:}
While RT-DETR was designed for generic object detection using bounding box queries, RowDetr adapts this paradigm for structured row detection by introducing polynomial proposals as queries. Instead of predicting bounding boxes, the network generates polynomial candidates from the encoded features by selecting the top M queries based on a logits head. Each selected query corresponds to a potential crop row in the image and is processed through an MLP to regress the polynomial coefficients that define its shape. To guide the model toward producing meaningful proposals, supervision is applied to the logits head. Following the work from \cite{detr}, gradient flow is detached at the decoder proposal stage to ensure training stability and mitigate the impact of noisy logits on downstream learning.
\subsubsection{PolySampler and Offset Network:}
\begin{figure}[ht]
    \includegraphics[width=1.1\textwidth]{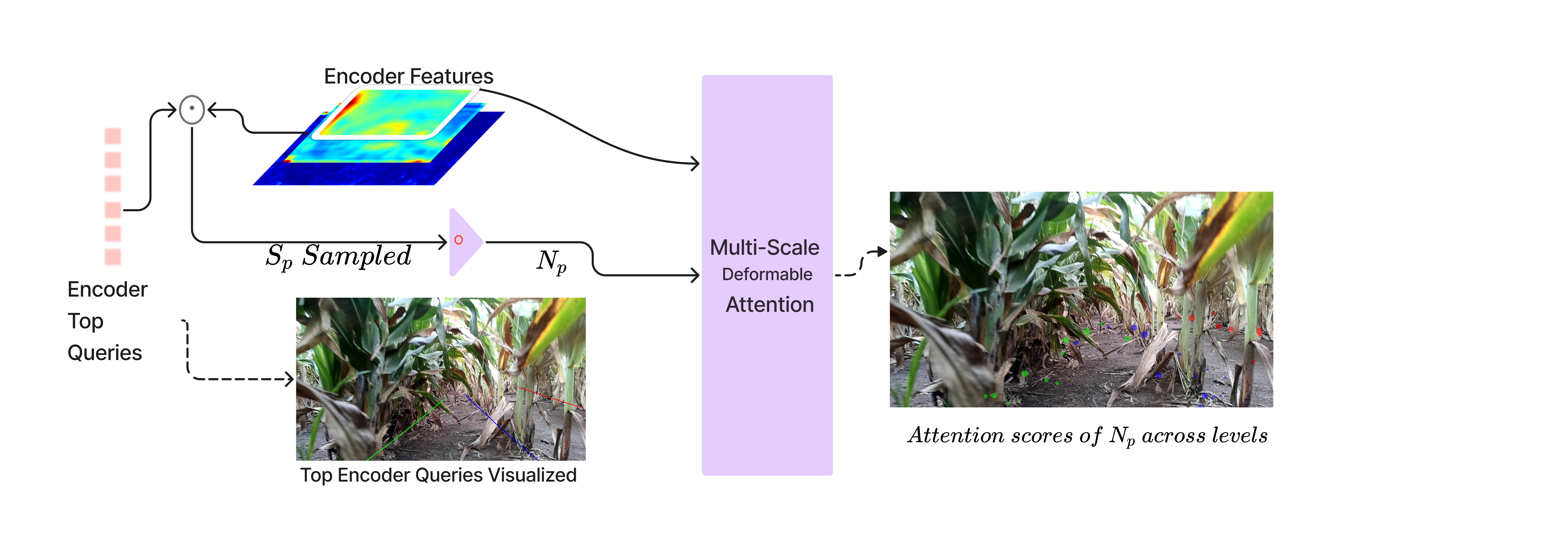}
    \caption{The top encoder queries (shown as red blocks) are projected onto the image space and sampled at equidistant intervals along polynomial trajectories denoted as \( S_p \). These sampled points capture structured features from the encoder feature map visualized in the heatmap. An offset network then refines these sampled points, producing \( N_p \), which are passed through the Multi-Scale Deformable Attention module. The right image visualizes the attention scores of \( N_p \) across feature levels for their respective polynomial (same color), with the size of the markers indicating attention magnitude. The colored curves (green, blue, red) represent top encoder proposals, displayed without any ranking order.}
    \label{fig:polysampler}
\end{figure}

The selected polynomial proposals are further refined using the PolySampler module and an offset network as shown in Fig. \ref{fig:polysampler}. In the PolySampler module, \( S_p \) points are sampled at equidistant values of \( \lambda \), capturing structured features along the polynomial curves. Further these sampled points embeddings are obtained using a grid-based strategy at its respective feature level.

The offset network refines these sampled points to align them better with ground truth rows. It consists of a two-layer Multi-Layer Perceptron (MLP) with an output dimension of \( 2 \times N_p \), where \( N_p \) (Eq. \ref{eq:offset_network}) represents the number of offset points. The final output of the MLP is adjusted using a scaled tanh activation function:

\begin{equation}
N_p = s \cdot \tanh(\text{MLP}(em)) + S_p
\label{eq:offset_network}
\end{equation}

where \( em \) represents the sampled embeddings at \( S_p \) point, and \( s \) (0.035) is a small scaling factor.

\subsubsection{Multi-Scale Deformable Attention:}
The refined feature points produced by the PolySampler and offset network are further processed by a Multi-Scale Deformable Attention module \cite{zhu2021deformabledetrdeformabletransformers}. This module enables the model to dynamically attend to spatially relevant regions across multiple feature scales, allowing it to effectively capture crop rows of varying shapes, orientations, and sizes. By focusing attention only on informative regions rather than the entire feature map, the model achieves both computational efficiency and improved localization. The decoder leverages these context-aware features to refine the polynomial coefficients, producing the final row predictions with enhanced accuracy and robustness.

\subsection{Training}
The training objective for RowDetr is to accurately predict polynomial curves representing crop rows and to correctly classify them. \textbf{PolyOptLoss}, is designed to minimize the discrepancy between the predicted and ground-truth curves.

\subsubsection{Loss Function: PolyOptLoss}

PolyOptLoss is an energy-based loss function designed to measure the geometric alignment between a predicted curve and a labeled ground-truth curve. Both the predicted curve, $\hat{C}$, and the ground-truth curve, $C$, are expressed as parametric polynomials: $\hat{u} = \hat{U}(\lambda), \hat{v} = \hat{V}(\lambda)$ and $u = U(\lambda), v = V(\lambda)$, respectively, with $\lambda \in [0, 1]$.

The Euclidean distance, $D(\lambda)$ (Eq. \ref{eq:distance}), between a point on the predicted curve and its corresponding point on the labeled curve is given by:
\begin{equation}
  D(\lambda) = \sqrt{(U(\lambda) - \hat{U}(\lambda))^2 + (V(\lambda) - \hat{V}(\lambda))^2}  
  \label{eq:distance}
\end{equation}

The energy at this point, $E(\lambda)$ (Eq. \ref{eq:enegry}), is defined as half the squared Euclidean distance:
\begin{equation}
  E(\lambda) = \frac{1}{2} D(\lambda)^2
  \label{eq:enegry}
\end{equation}

The total energy, or \textbf{PolyOptLoss} ($L_{poly}$, Eq. \ref{eq:polyoptloss}), is the integral of this energy function over the entire parameter space of the curve. Minimizing this integral drives the predicted polynomial to align with the ground-truth polynomial.
\begin{equation}
  L_{poly} = \int_0^1 E(\lambda) \, d\lambda
  \label{eq:polyoptloss}
\end{equation}

To compute the integral efficiently, an analytical solution was formulated to operate directly on the polynomial coefficients. This approach eliminates the need for numerical integration, enhancing computational performance. The details of this coefficient-based loss calculation are presented in Algorithm \ref{algo:polyoptloss}.

\begin{algorithm}
    \caption{PolyOptLoss: Energy-based Loss for Polynomial Alignment}
    \label{algo:polyoptloss}
    \begin{algorithmic}[1]
    \Require Predicted polynomial coefficients $\mathbf{P} \in \mathbb{R}^{B \times M \times D}$
    \Require Target polynomial coefficients $\mathbf{T} \in \mathbb{R}^{B \times 1 \times D}$
    \Ensure Energy-based loss $\mathcal{L} \in \mathbb{R}^{B \times M}$
    \State $d \gets D - 1$ \Comment{Degree of polynomial}
    \State $\mathbf{E} \gets \mathbf{P} - \mathbf{T}$ \Comment{Error tensor}
    \State $\mathbf{A}_{i,j} \gets \frac{1}{i+j+1}$ \Comment{Analytical integration matrix}
    \State $\mathcal{L} \gets \sum_{i,j=0}^d \mathbf{E}_{i} \mathbf{A}_{i,j} \mathbf{E}_{j}$ \Comment{Quadratic form for the integral}
    \Return $\mathcal{L}$
    \end{algorithmic}
\end{algorithm}

\subsubsection{Optimal Matching and Total Loss}

The final training objective combines the geometric alignment loss with a classification loss. For a given target row $T$ and a predicted row $P$, the total loss, $L^{TP}$ (Eq. \ref{eq:total_loss}), is defined as:

\begin{equation}
    L^{TP} = L_{poly}^{TP} + L_{cls}^{TP}
    \label{eq:total_loss}
\end{equation}

where $L_{poly}^{TP}$ is the PolyOptLoss applied to the polynomial coefficients and $L_{cls}^{TP}$ is the binary cross-entropy loss for classifying the predicted row as a valid detection.

To assign ground-truth rows to the most suitable predictions in a one-to-one manner, the \textbf{Hungarian algorithm} is employed. This serves as a critical component of the training strategy, ensuring consistent supervision. For $N$ target rows and $M$ predicted rows, the algorithm determines the optimal assignment $\sigma^*$ that minimizes the total matching cost, $\mathcal{C}(\sigma)$:
\begin{equation}
    \sigma^* = \arg \min_{\sigma}\mathcal{C}(\sigma) 
\end{equation}

Here, $\sigma$ is a permutation that maps each target row $T_i$ to a predicted row $P_j$. This approach guarantees that each ground-truth row is uniquely matched to the best-fitting prediction, which is crucial for stable training and accurate evaluation. This matching objective simplifies the training pipeline by eliminating the need for post-processing steps like Non-Maximum Suppression (NMS).
The total cost for a given permutationis given by:
\begin{equation}
    \mathcal{C}(\sigma) = \sum_{i=1}^{N} L^{T_{\sigma(i)} P_i}
\end{equation}

\subsection{Evaluation Metrics}
\label{sec:metrics}

The evaluation of end-to-end row detection models currently depends on specific detection tasks, such as segmentation-based approaches \cite{CRDLD} or line-detection metrics \cite{agronav}. However, these approaches highlight the need for metrics that are adaptable across different detection methods. To address this, metrics such as TuSimple \cite{tusimple} and LPD \cite{LPD} are adapted from lane detection metrics for evaluation and comparison of the final models. The following evaluation metrics were used to compare RowDetr with other end-to-end methods.

\begin{enumerate}

    \item \textbf{Mean Poly Distance (MPD):} It calculates the trapezoidal integral of the squared distance between predicted and ground truth polynomials, after matching predictions to targets based on minimum distance. Unmatched predictions (false positives) and missing targets (false negatives) incur a penalty corresponding to the maximum energy value (set to 2 in this case). The MPD is defined as:
    \begin{equation}
        MPD = \sum_{\lambda=0}^1 \left[(U(\lambda) - \hat{U}(\lambda))^2 + (V(\lambda) - \hat{V}(\lambda))^2 \right] \cdot 0.01,
    \end{equation}
    where \( U(\lambda) \) and \( V(\lambda) \) represent the ground truth polynomials, and \( \hat{U}(\lambda) \) and \( \hat{V}(\lambda) \) represent the predicted polynomials.

    \item \textbf{Accuracy Percent (AP):}
    This metric quantifies the proportion of correct predictions relative to the total number of predictions, and is expressed as:
    \begin{equation}
        AP = \frac{\text{No. of Correct Predictions}}{\text{Total No. of Predictions}}.
    \end{equation}

    \item \textbf{TuSimple Metrics:}
    The TuSimple benchmark introduced in \cite{tusimple} evaluates performance using three key metrics: accuracy (\( \text{Acc} \)), false positive rate (\( \text{FP} \)), and false negative rate (\( \text{FN} \)). For a predicted row to be considered a true positive, its accuracy, defined as:
    \begin{equation}
        \text{Acc}(P_j, L^*_j) = \frac{1}{|L^*_j|} \sum_{(x^*_{i,j}, y^*_{i,j}) \in L^*_j} 1\left[ |p_j(y^*_{i,j}) - x^*_{i,j}| < \tau_\text{acc} \right],
    \end{equation}
    The Acc must be greater than or equal to a predefined threshold \( \varepsilon \) to be considered as a true positive. Here:
    \begin{itemize}
        \item \( L^*_j \): Set of ground truth points for the \( j \)-th row.
        \item \( (x^*_{i,j}, y^*_{i,j}) \): Ground truth coordinates.
        \item \( p_j(y^*_{i,j}) \): Predicted \( x \)-coordinate for \( y^*_{i,j} \).
        \item \( \tau_\text{acc} \): Maximum allowed deviation (20 pixels in TuSimple benchmark).
        \item \( \varepsilon \): Accuracy threshold (set to 0.85).
    \end{itemize}
    The average values of \( \text{Acc} \), \( \text{FP} \), and \( \text{FN} \) are computed across all images, and the F1 scores are reported. This metric, however, can be sensitive to local errors as it considers all rows in the image, as noted by \cite{e2emetrics}.

    \item \textbf{Lane Position Deviation (LPD):}
    Introduced by \cite{LPD}, it is adapted to emphasize the ego lane and to assign higher weight to the points closer to the camera. Ego lane refers to the lane where the robot is positioned. In this context:
    \begin{itemize}
        \item The left boundary of the ego lane is the row closest to the center on the left.
        \item The right boundary is the row closest to the center on the robot's right.
    \end{itemize}
    This metric quantifies the deviation of predicted rows from their expected positions relative to the robot’s perspective, providing a robust evaluation for row detection tasks.

    \item \textbf{Latency:}
    It is measured in milliseconds (ms) and is calculated by the difference between the time the image is captured and the time the prediction is made. This metric shows the computational efficiency of the model. 
\end{enumerate}

\subsection{Experimental Setup:}
All experiments were performed on a workstation equipped with two NVIDIA RTX 4090 GPUs with FP32 precision and an AMD Ryzen Threadripper PRO 5955WX 16-core processor. The implementation was developed in Python 3.8 using PyTorch 2.1 and MMEngine \cite{mmengine2022}, running on Ubuntu 22.04. Trained models were subsequently deployed with TensorRT on the NVIDIA Jetson Orin AGX platform.

\subsection{Baseline Selection and Implementation Details}

The proposed RowDetr was compared with two recent end-to-end methods, including AgroNav \cite{agronav} and RowColAttention \cite{RowColRowDetection}. For compatibility with AgroNav’s format, the polyline labels were converted by fitting straight lines to the ego lane (defined in Section~\ref{sec:metrics}) and extending them across the image. RowColAttention was reimplemented with the same loss functions and hyperparameters as in the original study to ensure consistency. Lightweight and efficient backbones were selected for the RowDetr.

\section{Results \& Discussion}

Table~\ref{tab:results} provides an overview of the various evaluation metrics. The edge latency was calculated by deploying the models on a Jetson Orin AGX with INT8 precision, and the values are presented as a bar graph in Fig. \ref{fig:latencies}. Section \ref{sec:backbone_performance_appendix} provides the performance of the RowDetr model on various testsets.
\subsection{Training Speed Analysis}

AgroNav was trained until its F1-score plateaued, while RowDetr was trained for 750 epochs. RowColAttention, despite reaching convergence around 93 epochs, was trained for 300 epochs to maintain consistency with prior work. 

Training durations varied across methods: AgroNav required approximately 8 hours, RowDetr 6 hours, and RowColAttention 5 hours. Notably, RowDetr, despite handling complex row detection tasks and being trained for more epochs, exhibited a shorter training time per epoch.

\begin{table}[ht]
\centering
\caption{Performance Comparison of Row Detection Models: Metrics accompanied by a downward arrow $\downarrow$ indicate that lower values are preferable, whereas metrics accompanied by an upward arrow $\uparrow$ indicate that higher values are preferable.}
\resizebox{\textwidth}{!}{
\begin{tabular}{lcccccc}
\hline
Model & Latency ($\downarrow$)   & Param Count ($\downarrow$) & LPD ($\downarrow$) & TuSimple F1 ($\uparrow$) & TuSimple FPR ($\downarrow$) & TuSimple FNR ($\downarrow$) \\
\hline
RowDetr[efficientnet] &  9.11ms & \textbf{23M} & \textbf{0.405} &  0.734 & 0.393 & 0.044  \\
RowDetr[resnet18] & 6.7ms & 31 M & 0.421 & 0.736 & 0.391 & \textbf{0.043}   \\
RowDetr[regnetx\_008] & 9.7ms & 27M & 0.416 & 0.725 & 0.404 & 0.046   \\
RowDetr[resnet50] & 9.25ms & 44M & 0.413 &  \textbf{0.740} & \textbf{0.384} & 0.046    \\
\hline
Agronav & 18 ms  & NA  &  0.825 & NA & NA & NA \\
\hline
RowCol\cite{RowColRowDetection} & 14.16ms & 35M & 1.48 & 0.3191 & 0.8028 & 0.0400 \\
\hline
\end{tabular}
}

\label{tab:results}
\end{table}

\subsection{Accuracy of Row Detection}
Fig.s~\ref{fig:comp} and~\ref{fig:comp_fig2} compare row detection performance among RowDetr, RowColAttention, and AgroNav under varying levels of occlusion and row curvature. In the center image of Fig.~\ref{fig:comp}, RowDetr demonstrates strong robustness by accurately detecting row structures despite partial occlusions, whereas AgroNav fails entirely in these challenging conditions. RowColAttention does detect the rows but struggles considerably when small herbs are present, often creating boundary-like artifacts due to its row and column attention mechanisms. 

These qualitative findings align with the quantitative metrics in Table~\ref{tab:results}, where RowColAttention shows a high LPD of 1.48 and a low F1-score of 0.3191. In contrast, RowDetr maintains much lower LPD values (0.405-0.421) and achieves higher F1 scores (up to 0.740), indicating superior alignment and detection consistency. AgroNav performs comparably to RowDetr in structured fields when only straight lines are present, but its inability to model curved rows (illustrated in the rightmost image of Fig.~\ref{fig:comp_fig2}) is reflected by a higher LPD of 0.825. Overall, RowDetr's capacity to handle complex occlusions and curved row structures makes it more effective in real-world agricultural scenarios, where row curvature and dynamic clutter often affect detection accuracy.

\subsection{Performance Efficiency Analysis}
\begin{figure}[ht]
    \centering
    \includegraphics[width=\textwidth]{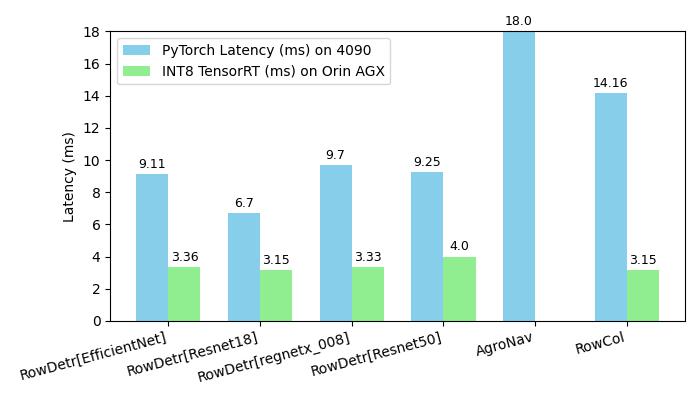}
    \caption{Latencies on RTX4090 and Jetson Orin AGX.}
    \label{fig:latencies}
\end{figure}

Fig. \ref{fig:latencies} shows the latency of the models that were compared. RowDetr demonstrates significantly lower latency compared to AgroNav and RowColAttention, with the RowDetr[ResNet18] variant achieving the lowest latency at \textbf{6.7\,ms}, nearly \textbf{3$\times$ faster} than AgroNav (18\,ms) and \textbf{2$\times$ faster} than RowColAttention (14.16\,ms). This efficiency is attributed to its optimized feature extraction and \emph{Poly Sampler and Offset Network} that reduces redundant computations in attention mechanism while maintaining real-time processing. For edge deployment, RowDetr[ResNet18] achieves comparable INT8 performance to RowColAttention on the Jetson Orin AGX, making it an attractive choice for resource-constrained environments. Regarding model complexity, RowDetr consistently employs fewer parameters across its variants, with the EfficientNet-based model having the smallest size at \textbf{23\,M parameters}, followed by RegNetX (27\,M) and ResNet18 (31\,M), which is nearly \textbf{34\% fewer parameters} than RowColAttention (35\,M). Despite these parameter reductions, RowDetr retains high computational efficiency and is well-suited for both high-performance and embedded systems.

In summary, these results confirm that RowDetr not only delivers higher accuracy but also operates more efficiently than both AgroNav and RowColAttention.

\subsection{On Variation of Backbones}
Among the tested backbones, ResNet50 delivered the highest accuracy in terms of TuSimple F1 Scores,  making it the best choice for scenarios prioritizing precision. RowDetr[efficientnet], with the lowest parameter count (23M), demonstrated competitive performance while being computationally efficient, making it ideal for resource-constrained systems.

\begin{figure}
  \centering
  \begin{subfigure}[t]{\textwidth}
    \centering
    \includegraphics[width=0.3\textwidth]{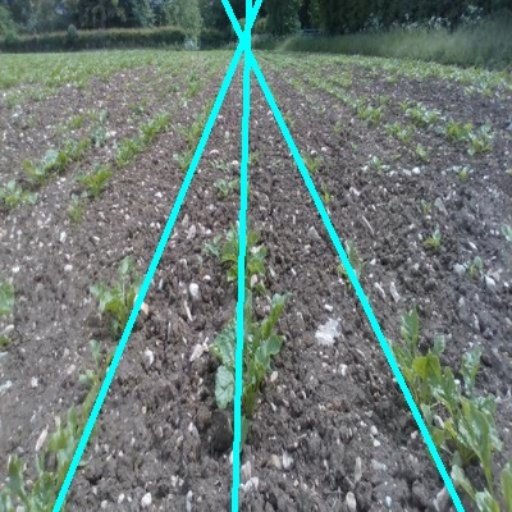}%
    \hfill
    \includegraphics[width=0.175\textwidth]{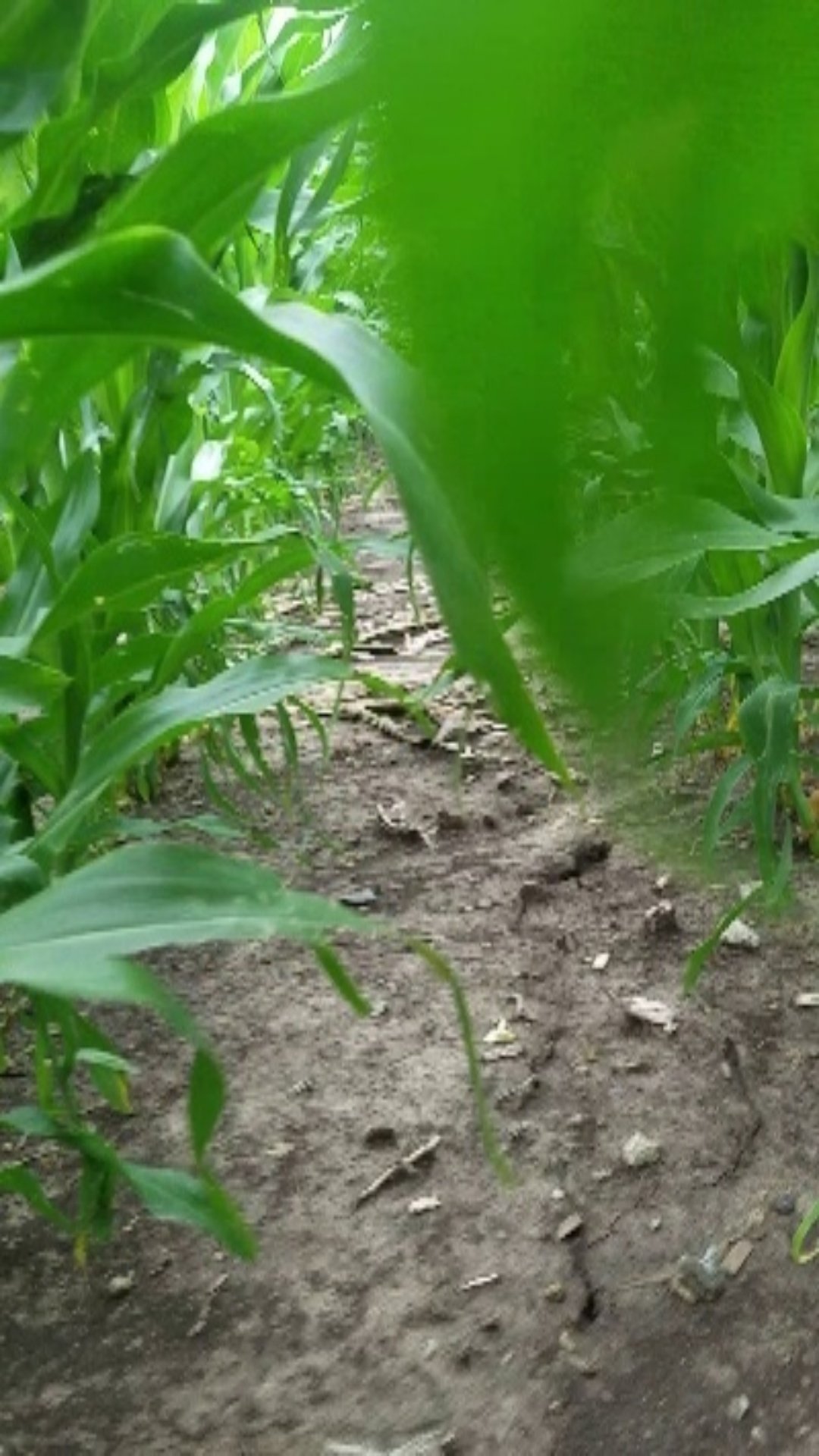}%
    \hfill
    \includegraphics[width=0.3\textwidth]{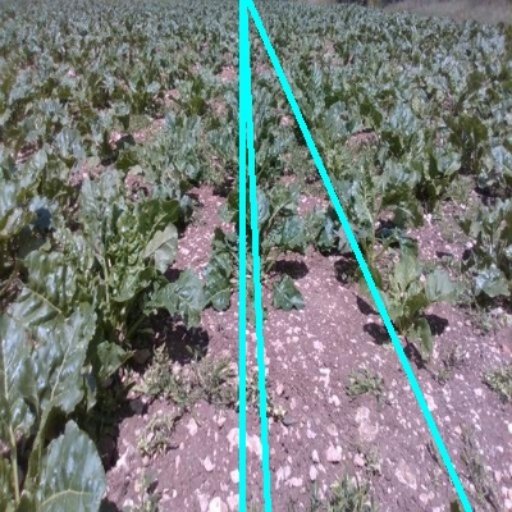}%
    \caption{Agronav results.}
    \label{fig:agronav1}
  \end{subfigure}
  
  \vspace{1em}  
  
  \begin{subfigure}[t]{\textwidth}
    \centering
    \includegraphics[width=0.3\textwidth]{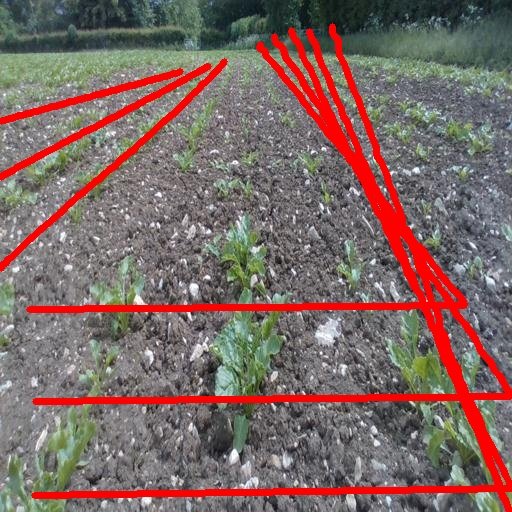}%
    \hfill
    \includegraphics[width=0.175\textwidth]{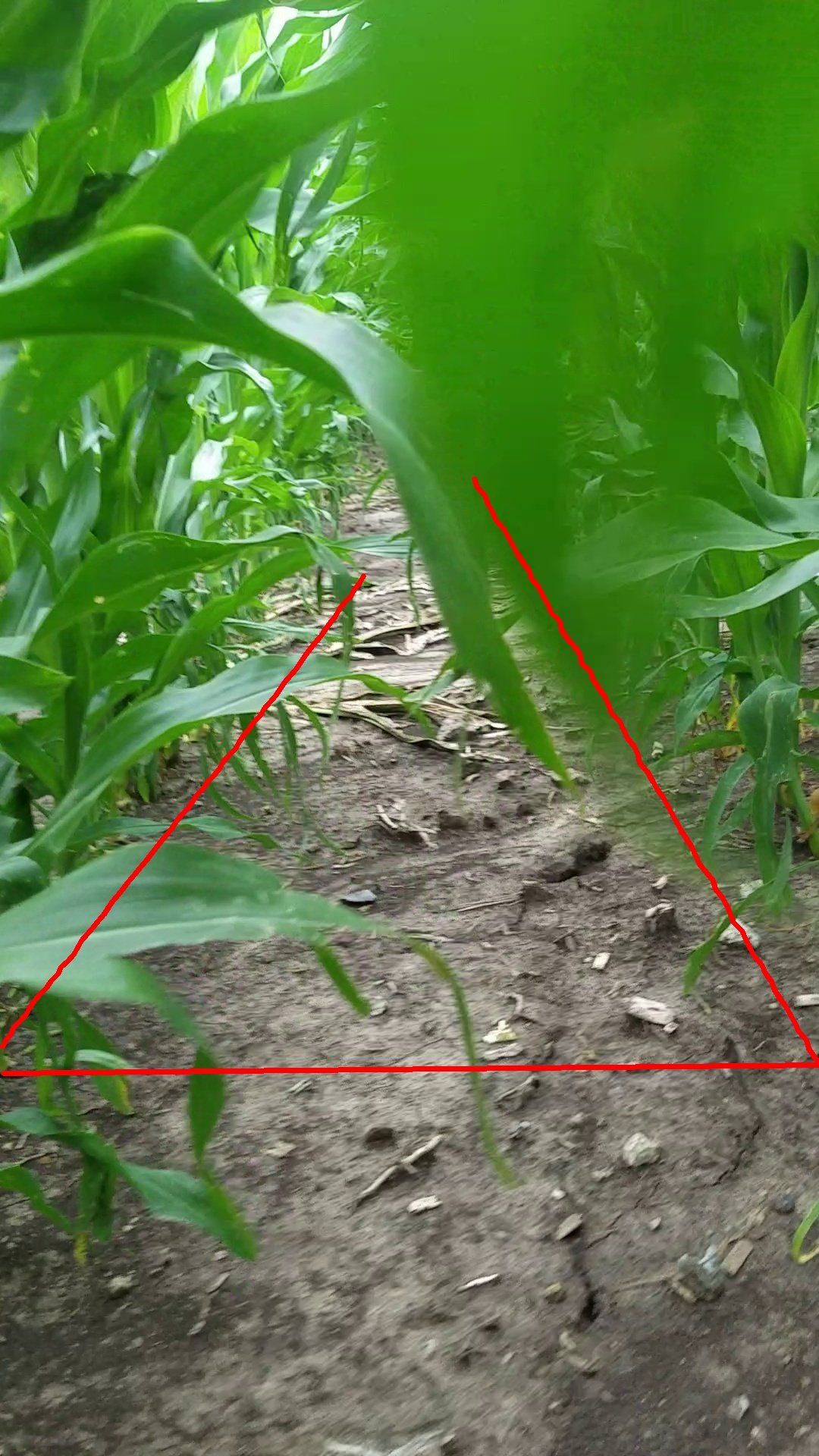}%
    \hfill
    \includegraphics[width=0.3\textwidth]{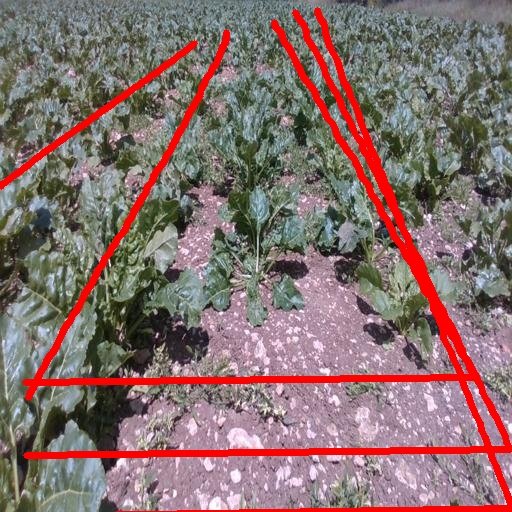}%
    \caption{RowCol results.}
    \label{fig:rowcol1}
  \end{subfigure}
  
  \vspace{1em}
  
  \begin{subfigure}[t]{\textwidth}
    \centering
    \includegraphics[width=0.3\textwidth]{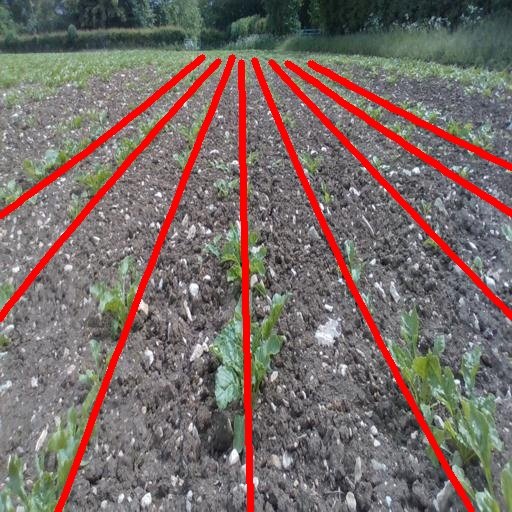}%
    \hfill
    \includegraphics[width=0.175\textwidth]{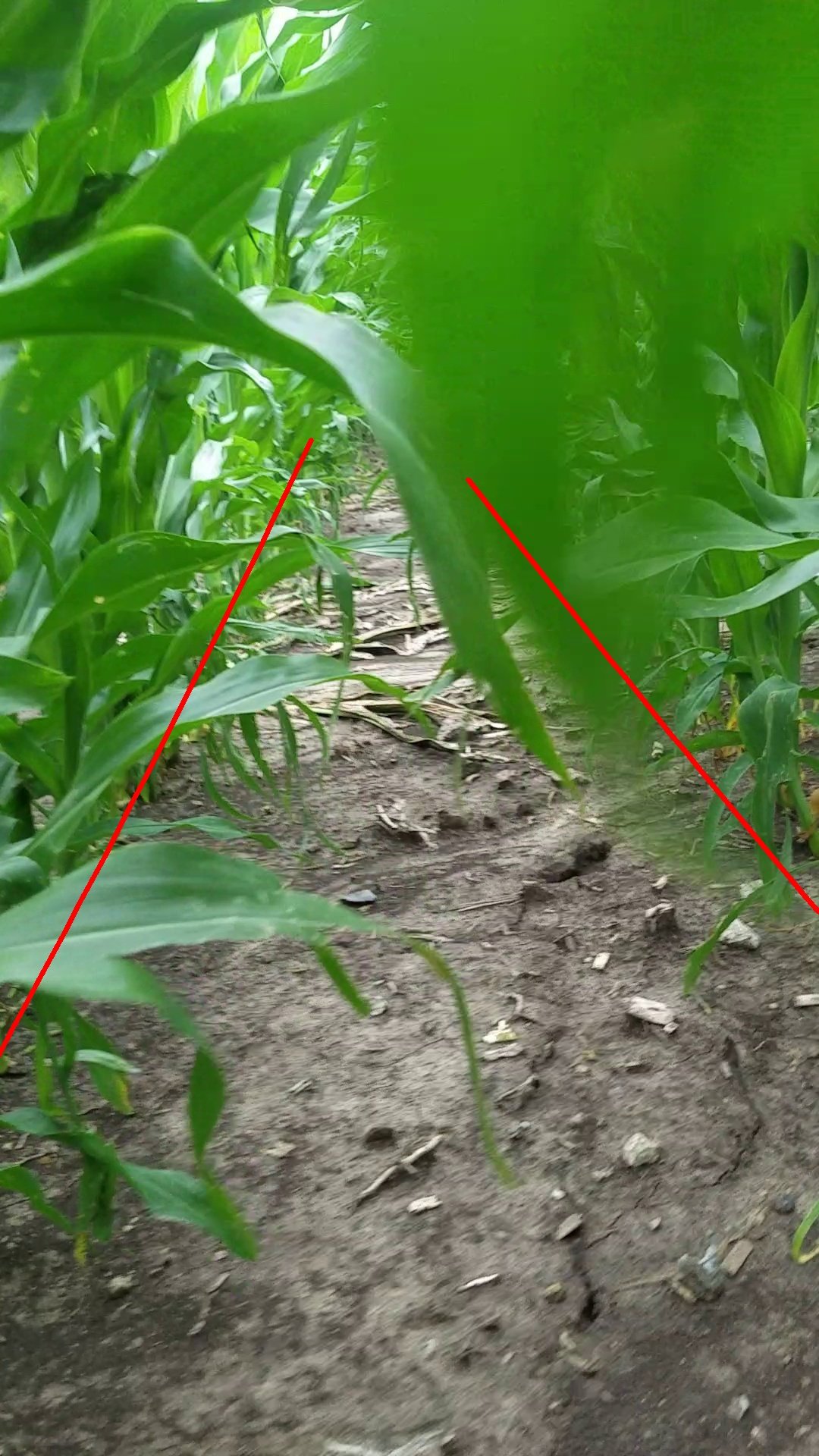}%
    \hfill
    \includegraphics[width=0.3\textwidth]{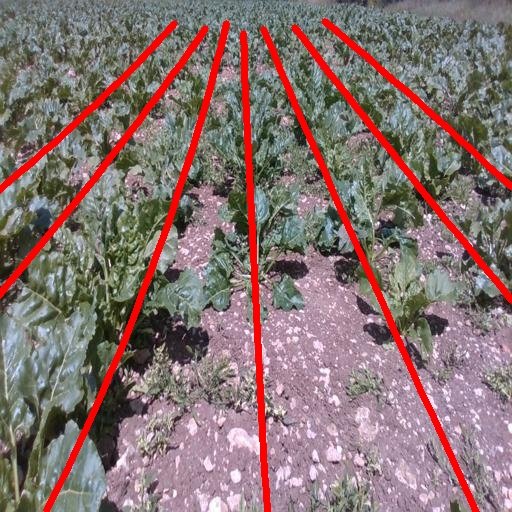}%
    \caption{RowDetr results.}
    \label{fig:rowdetr1}
  \end{subfigure}
  
  \caption{Comparison of results using the first three images for each category.}
  \label{fig:comp}
\end{figure}

\begin{figure}
  \centering
  \begin{subfigure}[t]{\textwidth}
    \centering
    \includegraphics[width=0.3\textwidth]{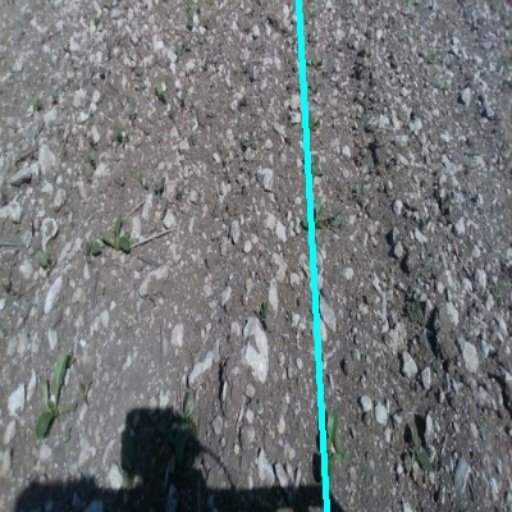}%
    \hfill
    \includegraphics[width=0.3\textwidth]{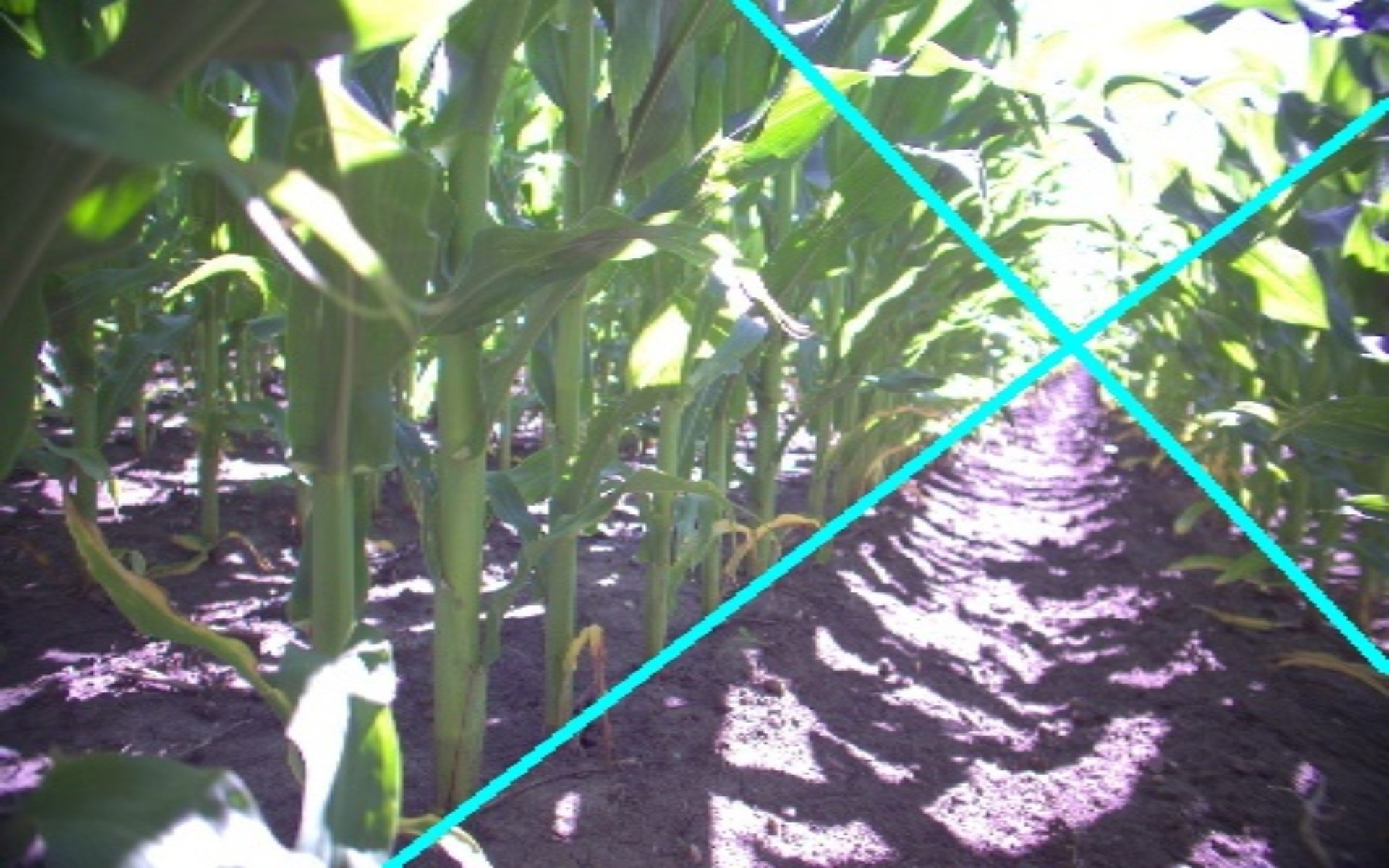}%
    \hfill
    \includegraphics[width=0.3\textwidth]{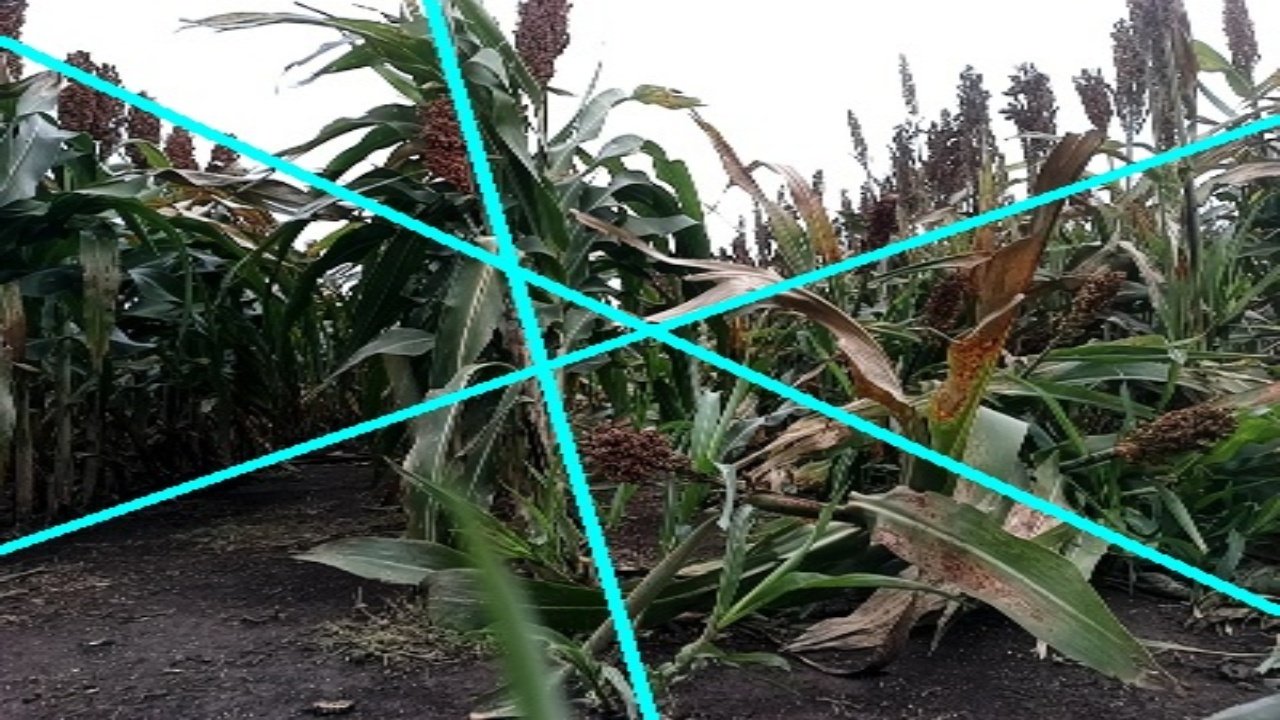}%
    \caption{Agronav results.}
    \label{fig:agronav2}
  \end{subfigure}
  
  \vspace{1em}
  
  \begin{subfigure}[t]{\textwidth}
    \centering
    \includegraphics[width=0.3\textwidth]{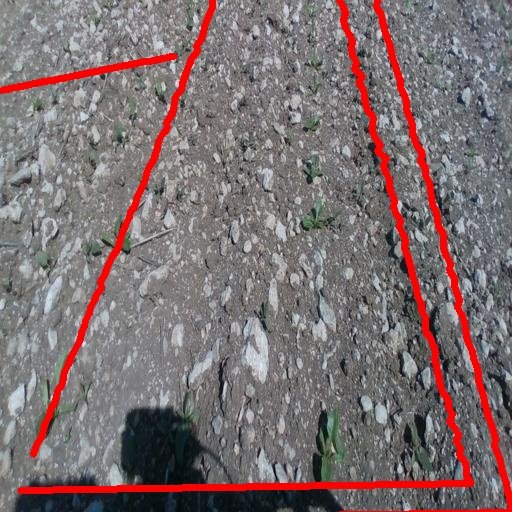}%
    \hfill
    \includegraphics[width=0.3\textwidth]{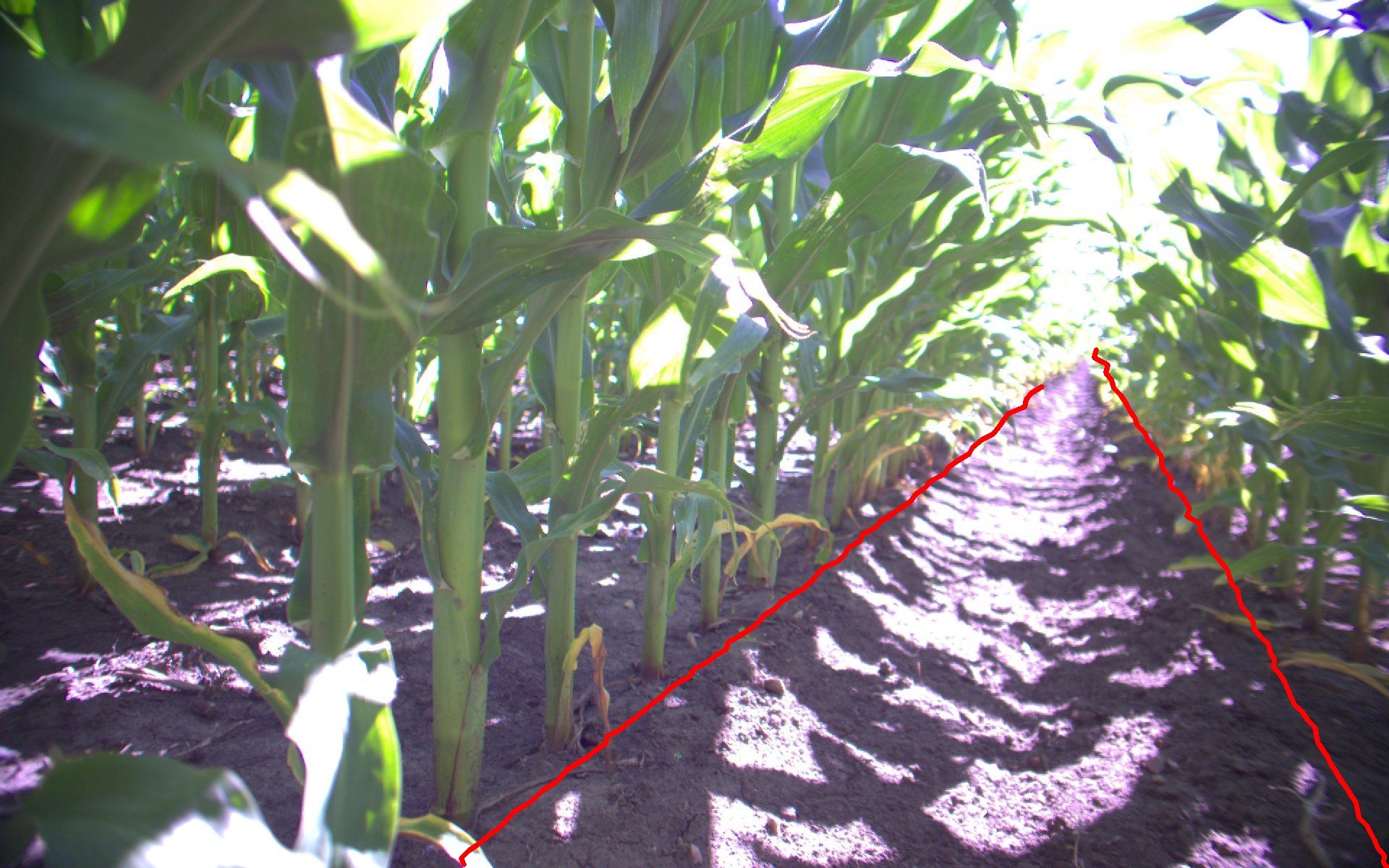}%
    \hfill
    \includegraphics[width=0.3\textwidth]{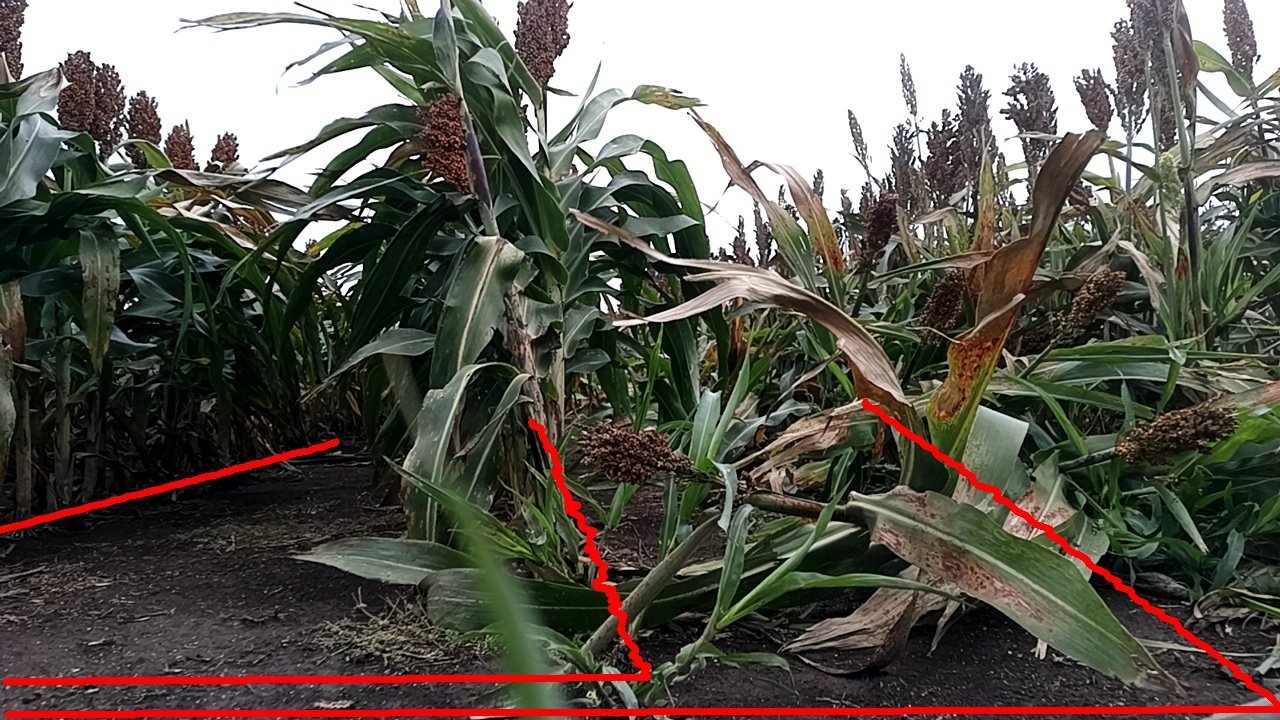}%
    \caption{RowCol results.}
    \label{fig:rowcol2}
  \end{subfigure}
  
  \vspace{1em}
  
  \begin{subfigure}[t]{\textwidth}
    \centering
    \includegraphics[width=0.3\textwidth]{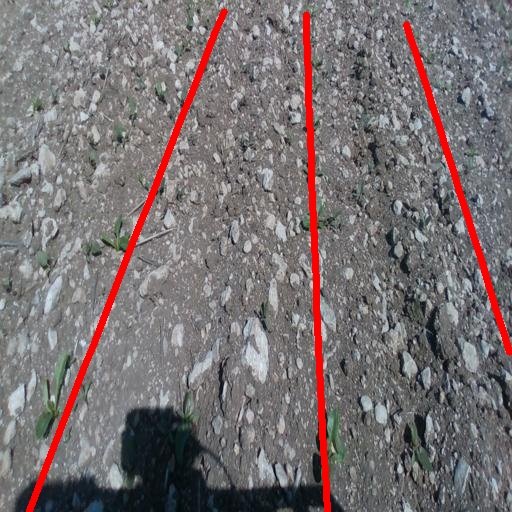}%
    \hfill
    \includegraphics[width=0.3\textwidth]{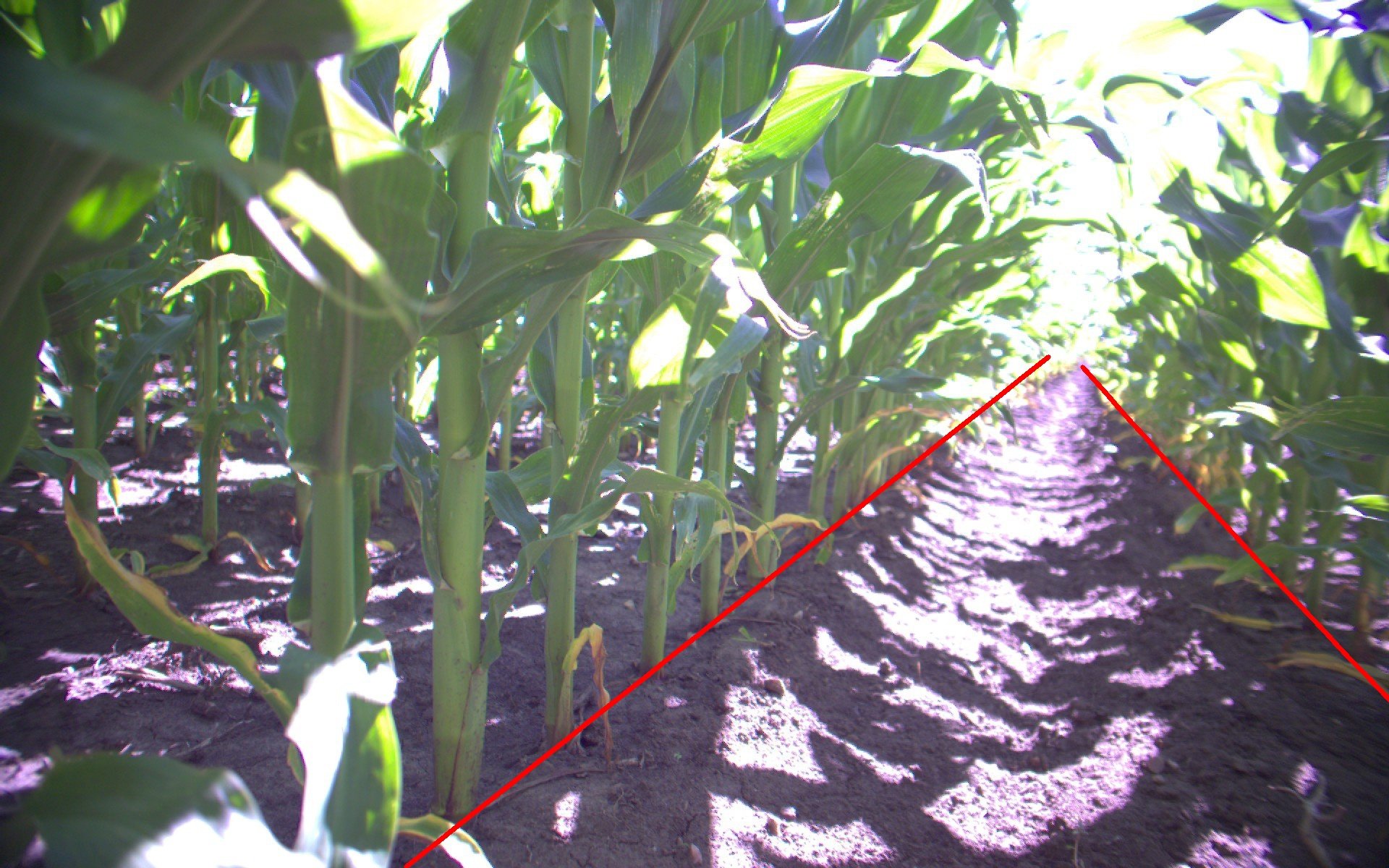}%
    \hfill
    \includegraphics[width=0.3\textwidth]{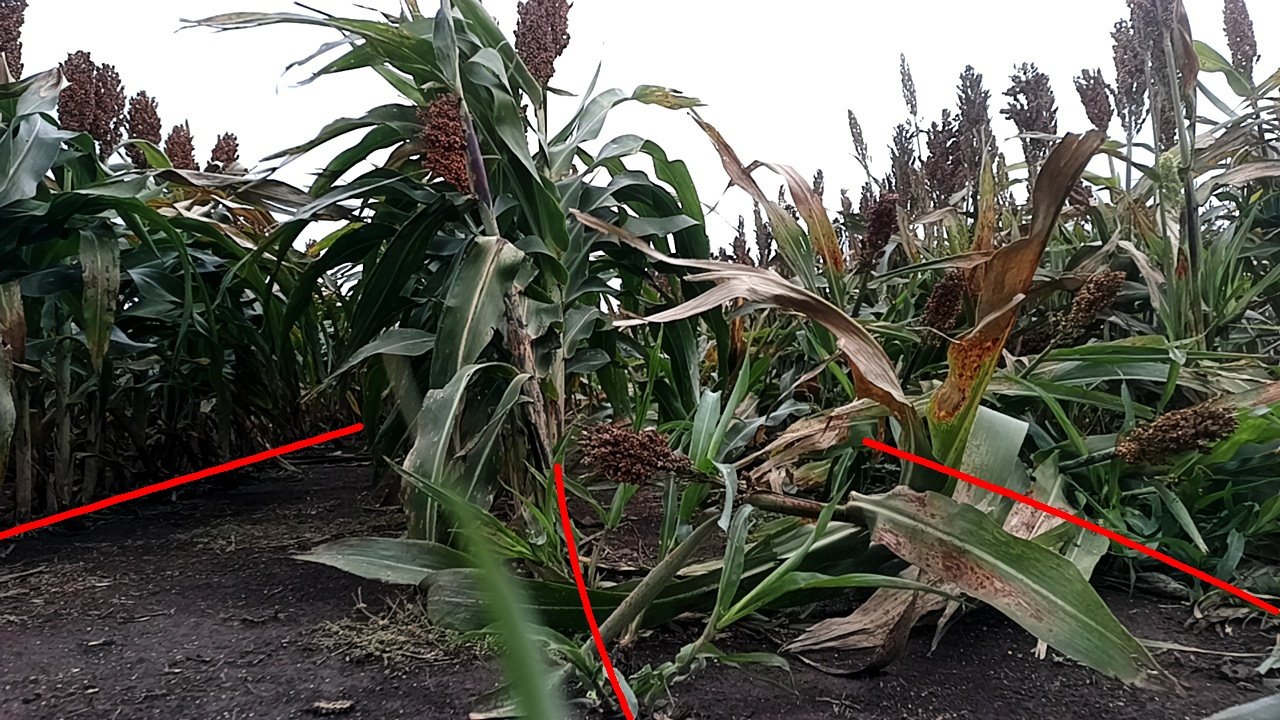}%
    \caption{RowDetr results.}
    \label{fig:rowdetr2}
  \end{subfigure}
  
  \caption{Continued Comparison.}
  \label{fig:comp_fig2}
\end{figure}

\subsection{Ablation Studies}
\subsubsection{Loss Study}

\begin{figure}[h]
    \centering
    \includegraphics[width=\textwidth]{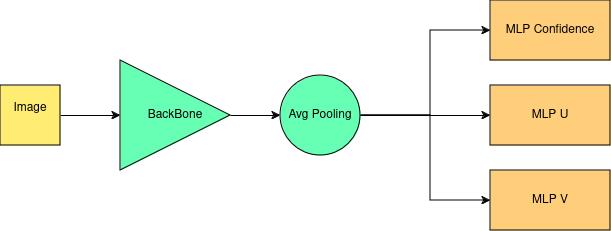}
    \caption{Model Architecture for Loss Study.}
    \label{fig:model-loss}
\end{figure}

\begin{figure}[t!]
    \centering
    \begin{subfigure}[t]{0.45\textwidth}
        \centering
        \includegraphics[width=0.95\textwidth]{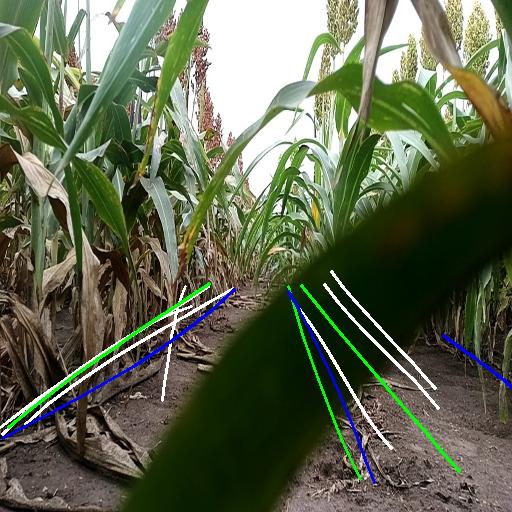}
        \caption{EfficientNet with Regression Loss.}
        \label{fig:eff_1}
    \end{subfigure}%
    ~
    \begin{subfigure}[t]{0.45\textwidth}
        \centering
        \includegraphics[width=0.95\textwidth]{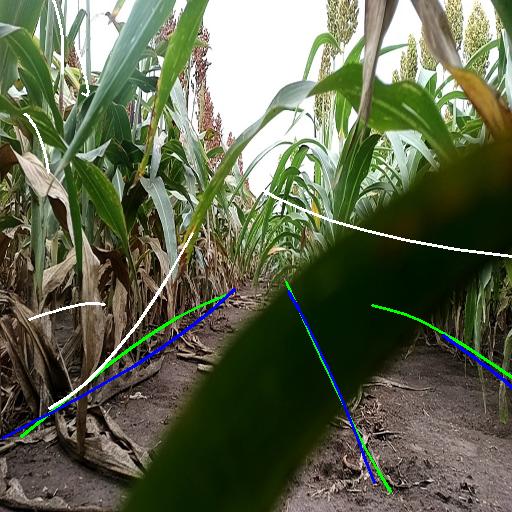}
        \caption{EfficientNet with PolyOptLoss.}
        \label{fig:eff_2}
    \end{subfigure}
    \vspace{0.5em}
    \begin{subfigure}[t]{0.45\textwidth}
        \centering
        \includegraphics[width=0.95\textwidth]{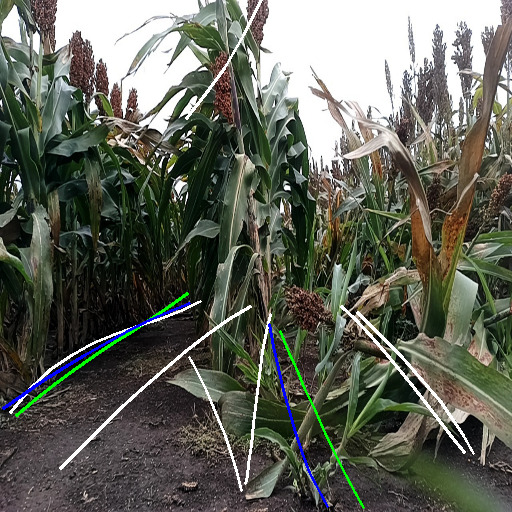}
        \caption{RegNet with Regression Loss.}
        \label{fig:reg_1}
    \end{subfigure}
    \begin{subfigure}[t]{0.45\textwidth}
        \centering
        \includegraphics[width=0.95\textwidth]{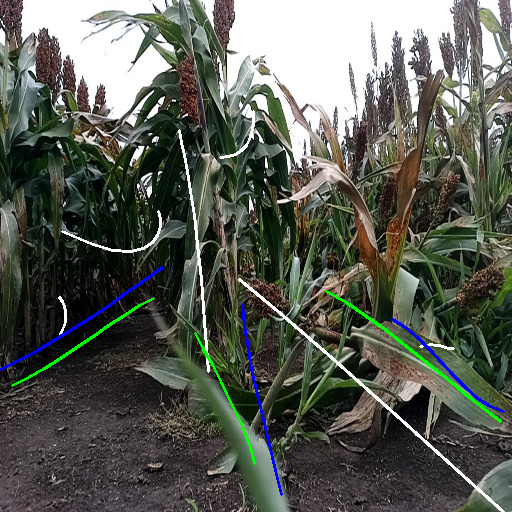}
        \caption{RegNet with PolyOptLoss.}
        \label{fig:reg_2}
    \end{subfigure}
    \caption{Sample detections in the validation set. 
    \footnotesize{White: Predictions with confidence \(< 0.5\); Blue: Labels; Green: Predictions with confidence \(> 0.5\).}}
    \label{fig:loss_study_imgs}
\end{figure}

To investigate the effectiveness of PolyOptLoss over a conventional regression-based loss (RMS error) in the presence of noisy labels, a subset of the sorghum dataset was selected and evaluated using two similar architectures. Both models employ a straightforward backbone followed by fully connected layers, as illustrated in Fig.~\ref{fig:model-loss}, to provide a controlled environment for assessing loss function impact without the complexity of transformer-based or attention-based components. The backbone network extracts high-level features from 512\(\times\)512 input images, and after experimentation with multiple backbones, EfficientNet~\cite{tan2019efficientnet} emerged as the best performer due to its Bi-FPNs, which enhance multi-scale feature fusion. Other networks such as ResNet-18~\cite{targ2016resnet} and RegNet~\cite{xu2022regnet} were also tested, but they did not match EfficientNet’s performance under this configuration. Following feature extraction, a global average pooling (GAP) layer condenses the feature maps into a single vector, which is then passed to three parallel MLPs: one predicts confidence scores via sigmoid activation, another outputs polynomial coefficients for the \(U\) component (horizontal alignment), and the third outputs coefficients for the \(V\) component (vertical alignment). Both polynomial MLPs use an ELU activation in the first layer and a linear activation in the final layer to ensure precise coefficient generation. The complete model architecture for the loss study is shown in Fig.~\ref{fig:model-loss}, and sample predictions are depicted in Fig.~\ref{fig:loss_study_imgs}. Table~\ref{tab:performance_metrics} summarizes the results, indicating that PolyOptLoss consistently outperforms Regression Loss across all tested backbones; for instance, EfficientNet reduces MPD from 0.3862 to 0.088 and boosts AP from 0.73496 to 0.9835 under PolyOptLoss, while RegNet and ResNet exhibit similar improvements. Qualitative assessments (Fig.~\ref{fig:loss_study_imgs}) reveal that Regression Loss often yields low-confidence predictions clustered near ground-truth rows and struggles under noisy conditions, whereas PolyOptLoss maintains higher-confidence predictions more closely aligned with annotated curves. These findings highlight PolyOptLoss’s superior capability in addressing occlusions, noisy labels, and other complexities characteristic of under-canopy detection tasks.

\begin{table}[h]
    \caption{Performance Metrics for the Loss Study.}
    \centering
    \begin{tabular}{cccc}
        \hline
        \textbf{Backbone} & \textbf{Loss} & \textbf{MPD} & \textbf{AP} \\
        \hline
        EfficientNet & Regression & 0.3862 & 0.73496 \\
        \textbf{EfficientNet} & \textbf{PolyOptLoss} & \textbf{0.088} & \textbf{0.9835} \\
        \hline
        RegNet & Regression & 0.400 & 0.7678 \\
        RegNet & \textbf{PolyOptLoss} & 0.097 & 0.9783 \\
        \hline
        ResNet & Regression & 0.769 & 0.7573 \\
        ResNet & \textbf{PolyOptLoss} & 0.1289 & 0.9657 \\
        \hline
    \end{tabular}
    \label{tab:performance_metrics}
\end{table}
\subsubsection{Sampling Points and Offset Points}
This section examines the influence of sampling points (\(S_p\)) and offset points (\(N_p\)) on model performance, with the polynomial degree fixed at 2 for consistency. Table~\ref{tab:spvsnp} illustrates that increasing \(S_p\) from 2 to 3 provides notable improvements in row alignment (lower LPD values), but further increases to 4 yield diminishing returns. Similarly, raising \(N_p\) from 1 to 2 refines predictions, while going to 3 shows negligible gains or slight degradation. The configuration \((S_p=3, N_p=2)\) emerges as the optimal balance, achieving the lowest LPD (0.416) and stable detection metrics. All tested settings maintain real-time performance under int8 quantization on the Orin-AGX platform, with only minor latency increases as \(N_p\) or \(S_p\) grows, thereby confirming the computational efficiency of the proposed approach.
\begin{table*}[ht]
\caption{Sampling Points and Offset Points Ablation Study.}
\centering
\begin{tabular}{llrrrrrr}
\toprule
 $S_p$ & $N_p$ & Mean LPD & TuSimple FNR & TuSimple FPR & TuSimple F1 \\
\midrule
2 & 1 & 0.441 & 0.049 & 0.427 & 0.705 \\
2 & 2 & 0.571 & 0.070 & 0.516 & 0.621 \\
2 & 3 & 0.419 & 0.035 & 0.422 & 0.715 \\
3 & 1 & 0.419 & 0.042 & \textbf{0.392} & \textbf{0.735} \\
\textbf{3} & \textbf{2} & \textbf{0.416} & \textbf{0.035} & 0.422 & 0.716 \\
3 & 3 & 0.417 & 0.048 & 0.417 & 0.714 \\
4 & 1 & 0.419 & 0.039 & 0.410 & 0.723 \\
4 & 2 & 0.452 & 0.042 & 0.418 & 0.716 \\
4 & 3 & 0.424 & 0.044 & 0.428 & 0.707 \\
\bottomrule
\end{tabular}
\label{tab:spvsnp}
\end{table*}
Table~\ref{tab:spvsnp} showcases the interplay between sampling points (\(S_p\)) and offset points (\(N_p\)), emphasizing their impact on row detection accuracy, false positive/negative rates, and computational efficiency. Increasing \(S_p\) from 2 to 3 improves the granularity of sampled embeddings, resulting in the lowest LPD (0.416) when \(N_p = 2\) and yielding the highest TuSimple F1 (0.735) when \(N_p = 1\). However, \(S_p = 4\) leads to diminished gains, consistent with expectations that three points suffice for second-degree polynomial fitting. Meanwhile, increasing \(N_p\) from 1 to 2 reduces FNR (e.g., from 0.042 to 0.035 at \(S_p = 3\)) and enhances LPD, though further increases to \(N_p = 3\) offer limited benefit. Consequently, the configuration \((S_p = 3, N_p = 2)\) provides the best overall trade-off, minimizing errors while preserving real-time inference (under 3.75\,ms latency on the Orin-AGX with int8 quantization). This balance between accuracy and speed, with minimal overhead from additional sampling or offsets, underscores the method’s suitability for under-canopy row detection tasks.

\subsection{Role of Encoder and Decoder in Row Detection}
\begin{figure}[t!]
    \centering
    \begin{subfigure}[t]{0.5\textwidth}
        \centering
        \includegraphics[width=0.95\textwidth]{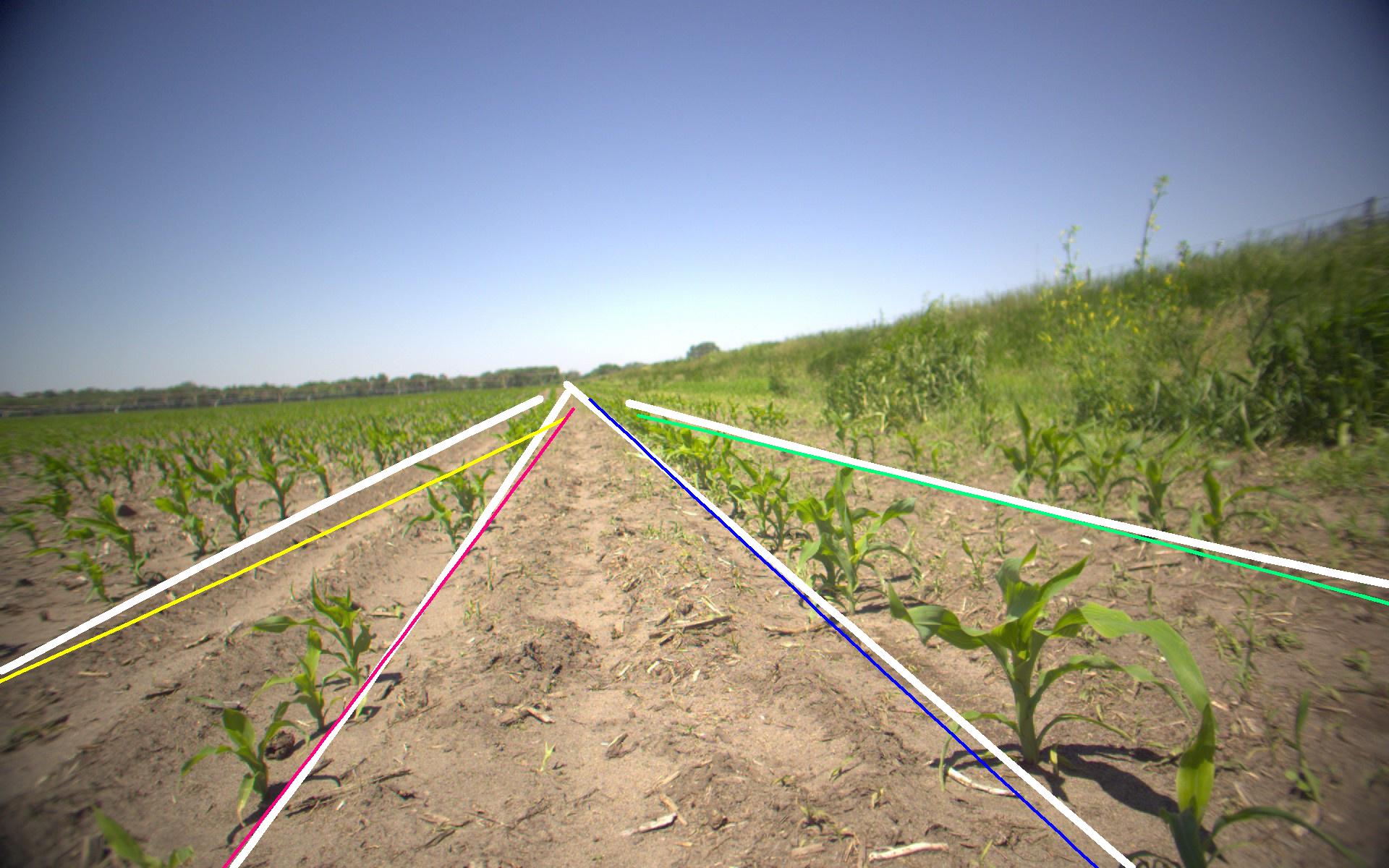}
        \caption{}
        \label{fig:encoder_out1}
    \end{subfigure}%
    ~ 
    \begin{subfigure}[t]{0.5\textwidth}
        \centering
        \includegraphics[width=0.95\textwidth]{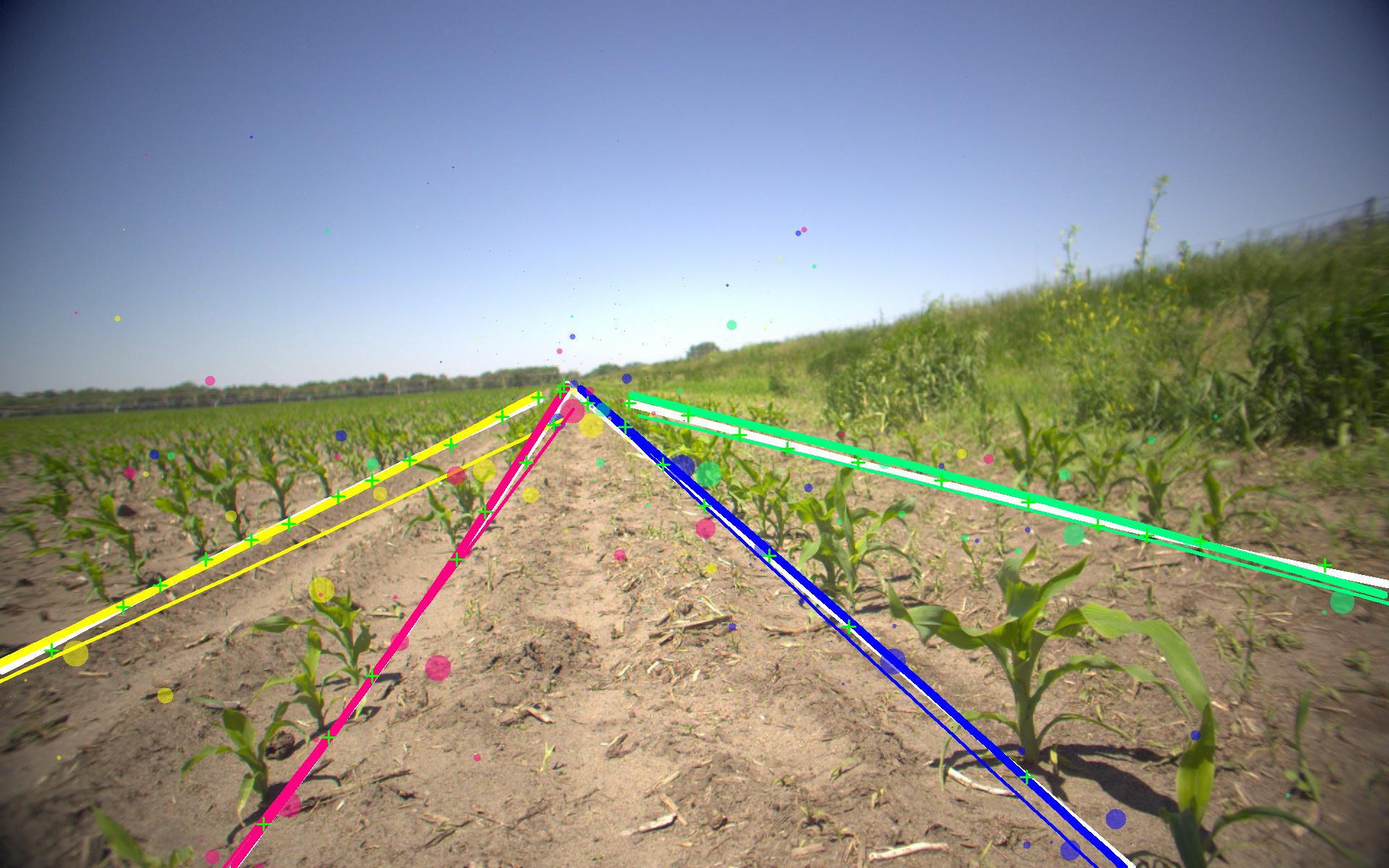}
        \caption{}
        \label{fig:decoder_out1}
    \end{subfigure}
    \hfill
    \begin{subfigure}[t]{0.5\textwidth}
        \centering
        \includegraphics[width=0.95\textwidth]{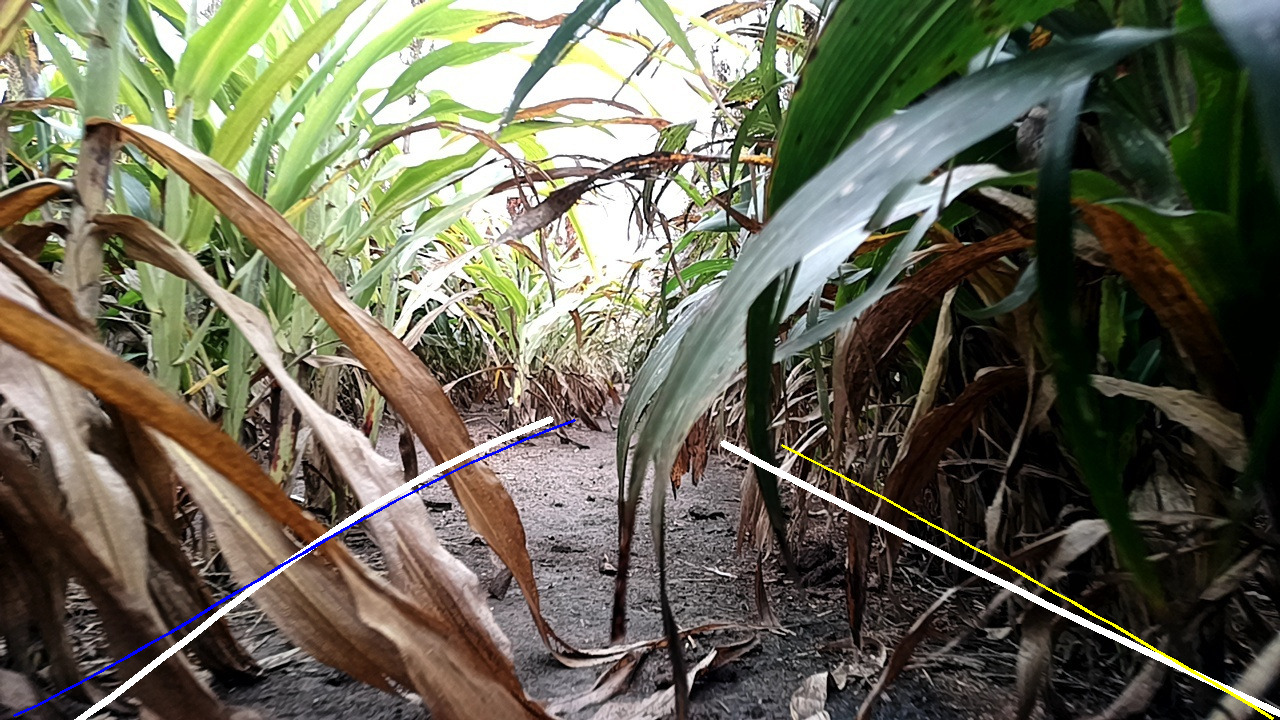}
        \caption{}
        \label{fig:encoder_out3}
    \end{subfigure}%
    ~ 
    \begin{subfigure}[t]{0.5\textwidth}
        \centering
        \includegraphics[width=0.95\textwidth]{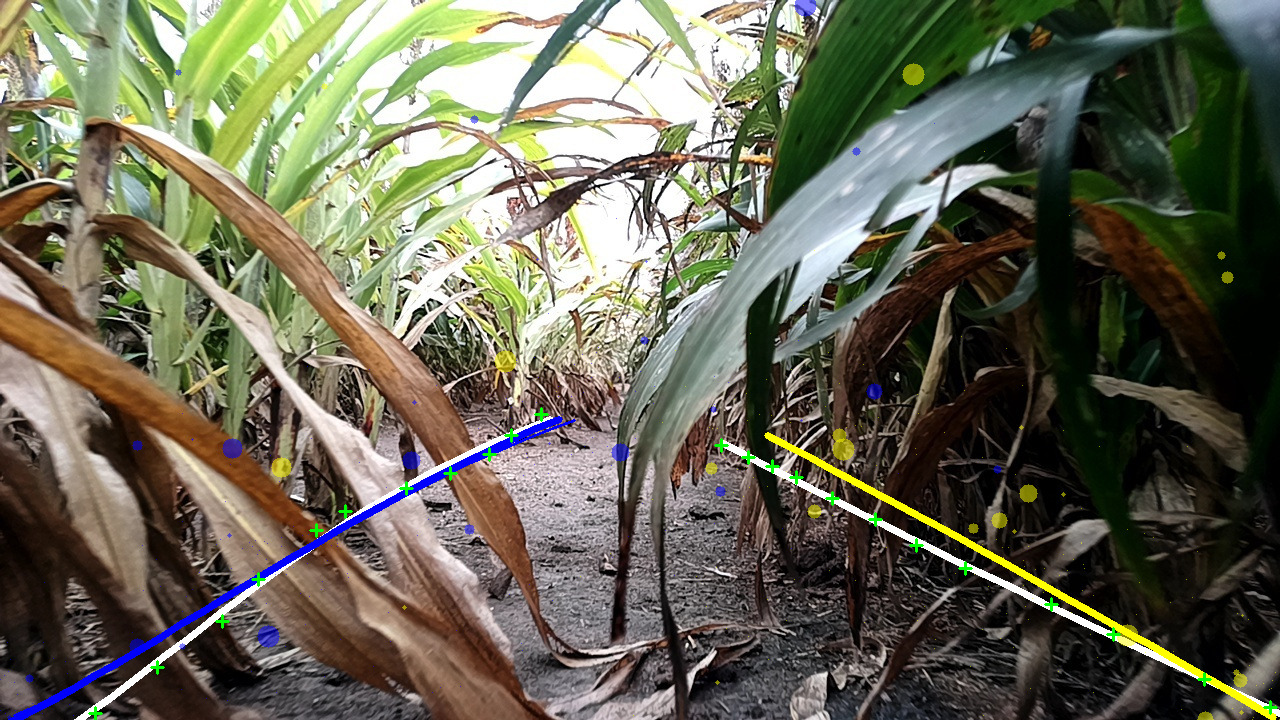}
        \caption{}
        \label{fig:decoder_out3}
    \end{subfigure}
    \hfill
    \begin{subfigure}[t]{0.5\textwidth}
        \centering
        \includegraphics[width=0.95\textwidth]{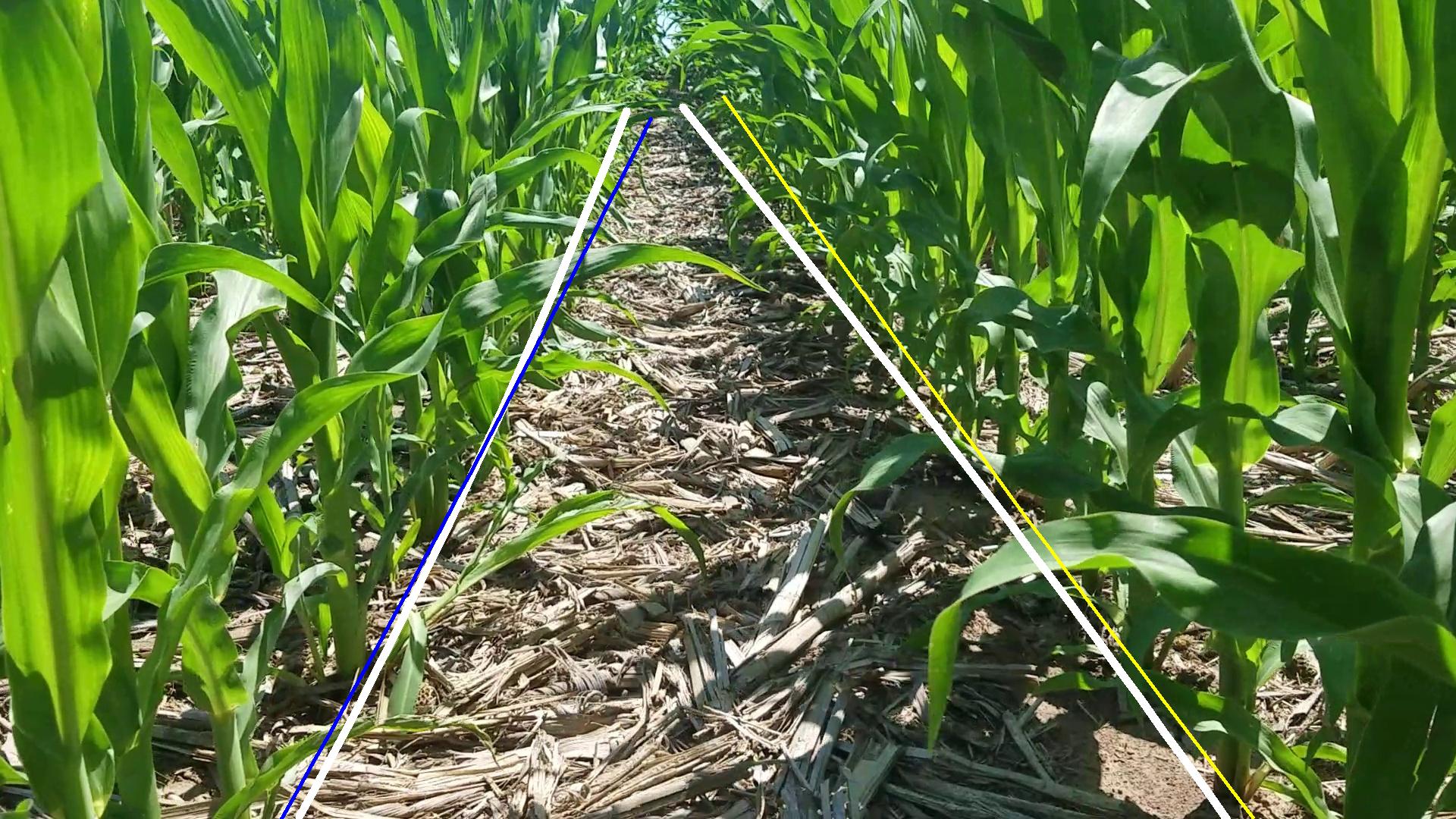}
        \caption{}
        \label{fig:encoder_out6}
    \end{subfigure}%
    ~ 
    \begin{subfigure}[t]{0.5\textwidth}
        \centering
        \includegraphics[width=0.95\textwidth]{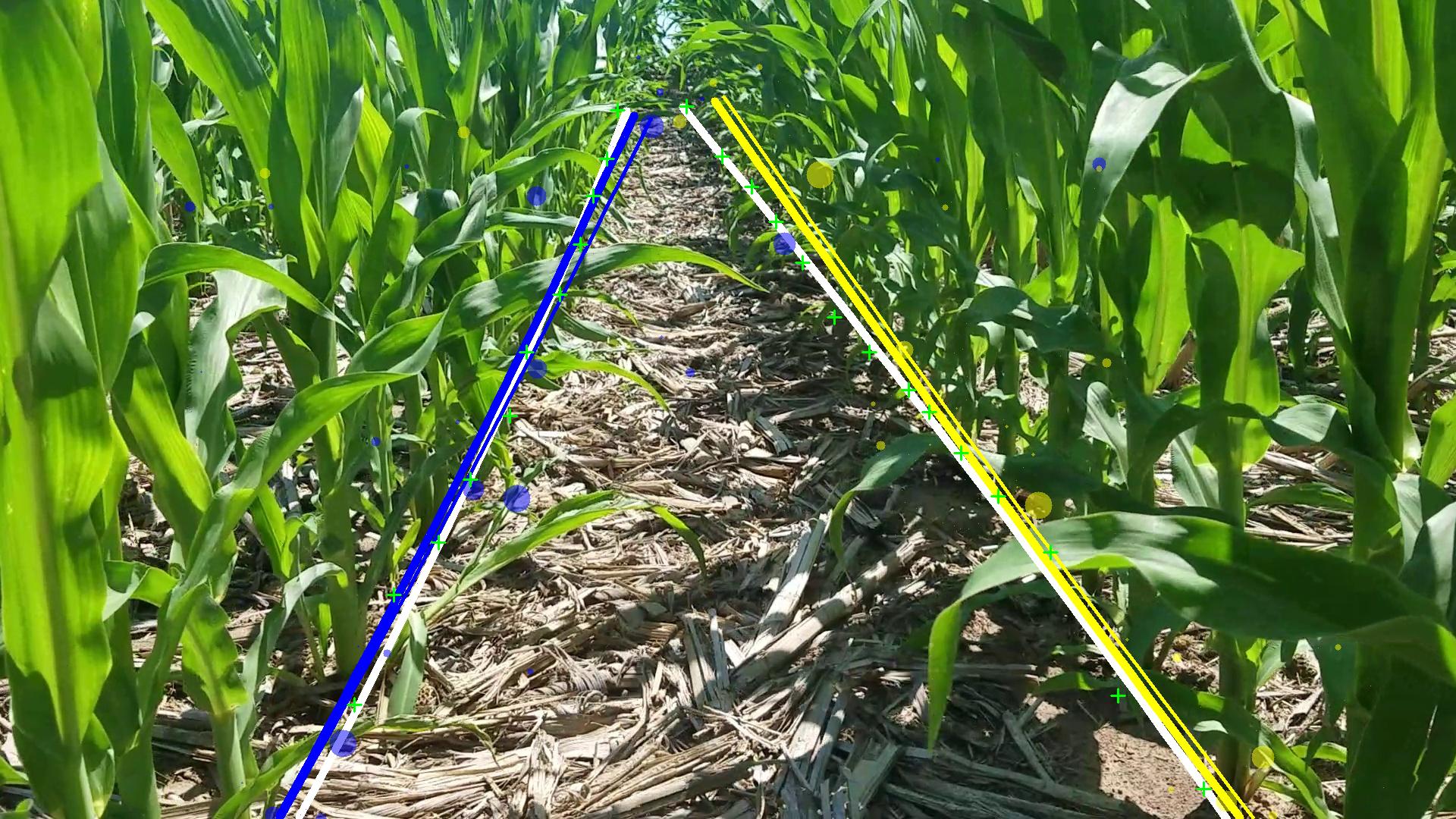}
        \caption{}
        \label{fig:decoder_out6}
    \end{subfigure}
    \hfill
    \begin{subfigure}[t]{0.5\textwidth}
        \centering
        \includegraphics[width=0.95\textwidth]{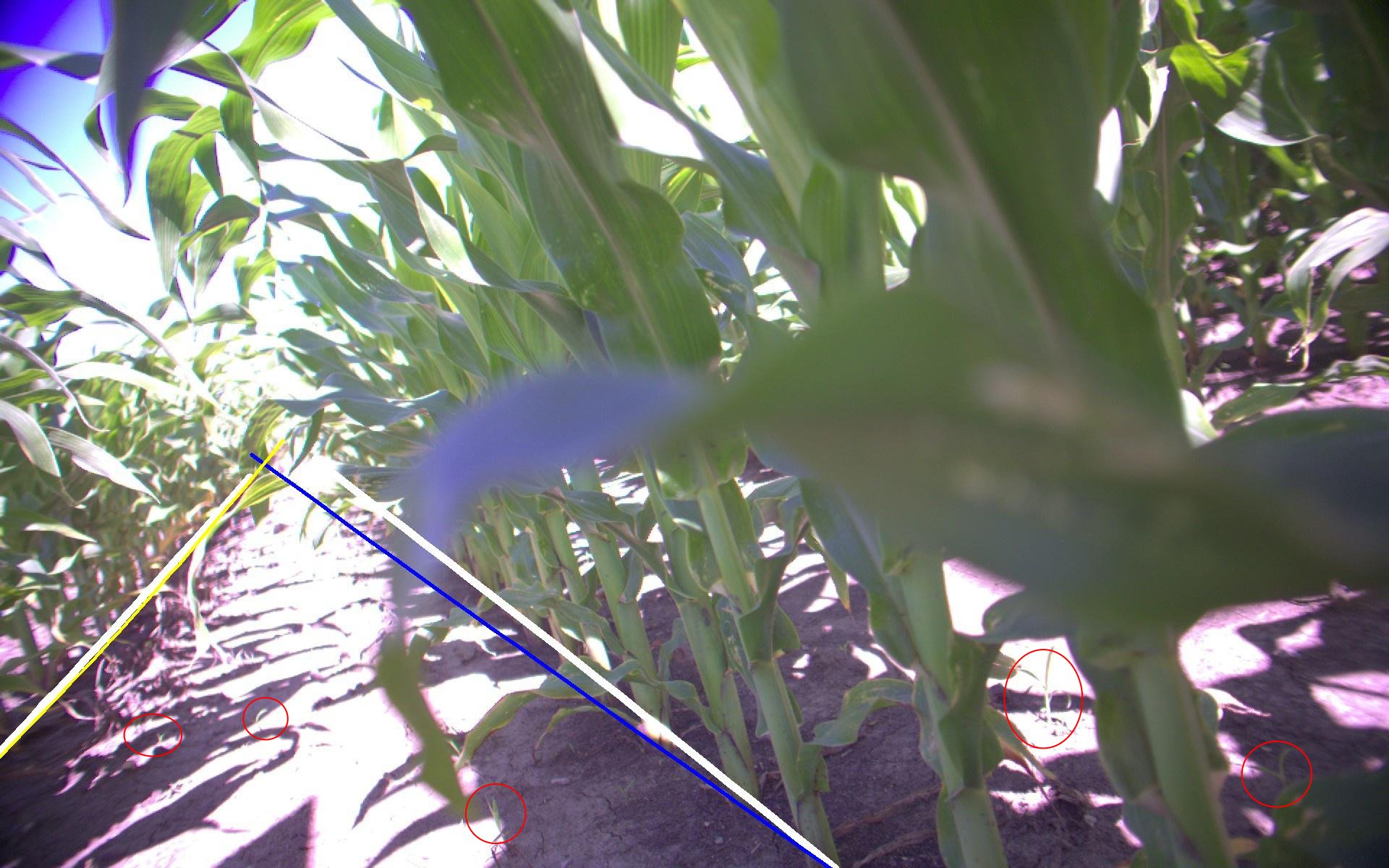}
        \caption{}
        \label{fig:encoder_out15}
    \end{subfigure}%
    ~ 
    \begin{subfigure}[t]{0.5\textwidth}
        \centering
        \includegraphics[width=0.95\textwidth]{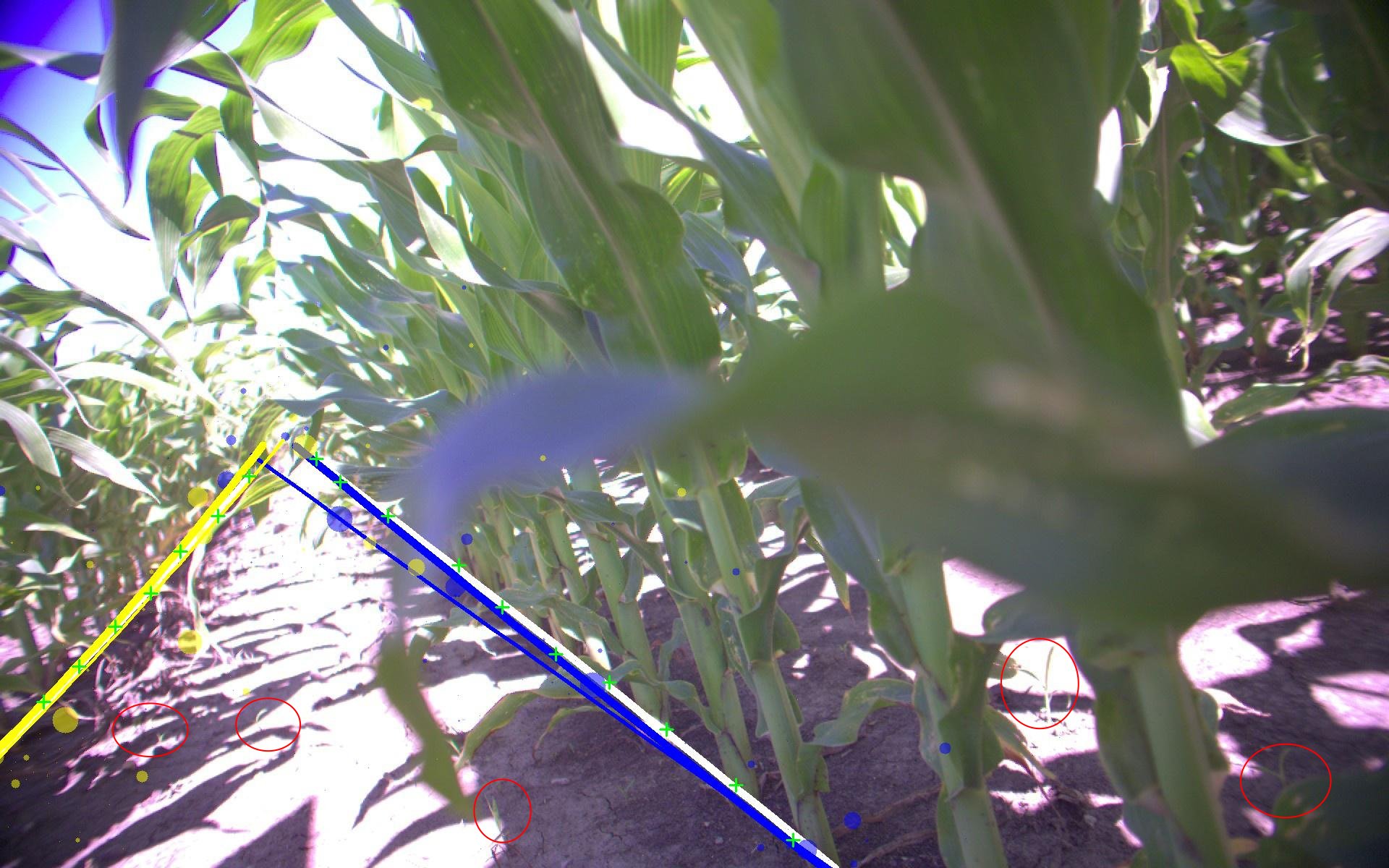}
        \caption{}
        \label{fig:decoder_out15}
    \end{subfigure}

    \hspace{5cm}
    \caption{Encoder (left) and decoder (right) outputs overlaid on ground-truth labels (white lines). Thin colored curves represent encoder predictions along polynomial proposals; thicker curves show decoder outputs after refinement. Attention score markers for sampled points \( N_p \) are also visualized, with larger markers indicating higher weights.}
    \label{fig:enc_decoder}

\end{figure}

 Figure~\ref{fig:enc_decoder} illustrates the outputs of the encoder and decoder stages in the proposed RowDetr. The left column corresponds to the encoder’s raw polynomial proposals, while the right column presents the refined detections generated by the decoder. To maintain consistency, proposals corresponding to the same row are assigned the same color across encoder and decoder outputs. Attention points are also visualized, where the size of each point is proportional to its attention score, thereby highlighting the regions most influential in shaping the final row proposal. 
 
 The encoder serves as the initial stage of the detection pipeline, producing polynomial proposals based on local image cues. However, because it operates with only partial global context, the encoder is prone to producing errors in challenging conditions. For example, as shown in Fig.~\ref{fig:encoder_out1}, the yellow proposal drifts into the trajectory of the pink proposal, effectively merging two distinct rows into a single structure. These types of misalignments are corrected in the decoder stage, which leverages both the polynomial proposals and the sampler network to enforce spatial consistency, as seen in Fig.~\ref{fig:decoder_out1}.
 
 A similar trend can be observed in Figs.~\ref{fig:encoder_out6} and \ref{fig:decoder_out6}, where encoder predictions incorrectly extend into the navigable path. Even under noisy conditions with plant residues, the decoder is able to refine these proposals, aligning them more closely with the true row boundaries. This demonstrates the decoder’s ability to use contextual cues and global consistency to suppress encoder errors that arise from occlusions or ambiguous textures.
 
 In scenes with dense crops, curved rows, or gaps between stalks (e.g., Fig.~\ref{fig:encoder_out3}), annotation becomes particularly challenging since stalk visibility is limited. This occasionally leads to errors in the ground-truth labels themselves, such as the white lines on the right side of the image. Despite these inconsistencies, the decoder demonstrates robustness by interpreting the overall scene context and adjusting the encoder’s raw proposals into coherent and valid row predictions. 
 
 Overall, the encoder can be viewed as a high-recall proposal generator, capable of quickly hypothesizing multiple potential row candidates. The decoder, in contrast, acts as a precision module that integrates global scene context, attention-weighted sampling, and structural constraints to refine these proposals into accurate and reliable row detections. This complementary design ensures both flexibility and robustness, enabling the system to perform reliably under diverse field conditions ranging from early-stage sparse crops to late-stage dense canopies.

\clearpage
\section{Conclusion}
This study addresses the challenges of Row Detection in dense agricultural environments by introducing a robust framework. The proposed contributions encompass key advancements in dataset curation, loss metric design, and model architecture, yielding significant improvements in performance and efficiency.

\begin{itemize}
    \item \textbf{PolyOptLoss:} The introduction of the \textit{PolyOptLoss} demonstrates a marked improvement over traditional loss functions, such as Regression Loss. By leveraging geometric information and minimizing alignment errors, PolyOptLoss consistently outperformed Regression Loss across various backbones, particularly in the presence of noisy labels and occlusions.
    \item \textbf{RowDetr Architecture:} The proposed RowDetr architecture establishes a new benchmark for row detection in dense and curved environments. Its capability to maintain high accuracy and robustness across a range of backbones underscores its flexibility and scalability. Additionally, its compatibility with optimization techniques like TensorRT ensures low-latency operation, making it suitable for real-time agricultural applications.
    \item \textbf{Ablation Studies:} Experiments on sampling points and offset points reveal that the optimal configuration (\( S_p = 3 \), \( N_p = 2 \)) strikes the best balance between computational efficiency and detection accuracy. These results highlight the importance of carefully tuning these parameters to maximize performance.
    \item \textbf{Comparative Evaluations:} Comparative assessments against AgroNav and RowColAttention demonstrate that RowDetr achieves superior performance across all metrics, including accuracy, false positive/negative rates, and latency. These findings establish RowDetr as a state-of-the-art framework for end-to-end row detection.
\end{itemize}

\paragraph{Limitations:}
Although RowDetr demonstrates strong performance in detecting rows within dense environments, it faces limitations in highly unstructured settings, such as forests, where the assumption of smooth curves does not hold. Furthermore, while the framework is adaptable to UAVs and diverse crop types given its reliance solely on images and annotations, the current dataset is restricted to corn and sorghum and has yet to be validated on other crop varieties.
\paragraph{Impact and Future Work:}
The outcomes of this study provide a foundation for developing more reliable and efficient autonomous systems tailored to agricultural applications. By addressing key limitations of existing methods such as ensuring low latency and improving robustness in noisy environments—the proposed framework demonstrates its potential for deployment in real-world scenarios, including GPS-denied conditions and densely planted fields.

Future work will focus on expanding the framework's generalizability to a wider range of crops and field conditions. Incorporating multi-modal sensor data (e.g., IR or thermal imaging). Additionally, leveraging self-supervised or semi-supervised learning techniques can reduce reliance on large labeled datasets, thereby increasing the practicality and scope of the approach. Research will also explore the development of navigation systems based on RowDetr detections, integrating them with the broader framework to further extend its functionality in autonomous agricultural operations.\section{Acknowledgments}
Funding for this project was provided by USDA-NIFA under the NSF National Robotics Initiative Program, Award \# 2019-67021-28995. We extend our gratitude to Zack Knust and Revanth Babu Raavi for their assistance with labeling, and to Ben Vail for his help in training data collection. We also wish to thank Jose Mateus Raitz and Niranjan Piya for their contributions to the robot’s design.

\section*{Code and Dataset Availability}

The source code for training and evaluating \textit{RowDetr}, along with configuration files, pretrained models, and sample data, is publicly available at: \url{https://github.com/r4hul77/RowDetr-v2}. The curated row detection datasets, including Corn V4–V6 and Sorghum sequences, will be released under an academic use license upon acceptance to support reproducibility and further research.

\FloatBarrier
\bibliographystyle{elsarticle-harv} 
\bibliography{example}

\begin{thebibliography}{44}
\expandafter\ifx\csname natexlab\endcsname\relax\def\natexlab#1{#1}\fi
\providecommand{\url}[1]{\texttt{#1}}
\providecommand{\href}[2]{#2}
\providecommand{\path}[1]{#1}
\providecommand{\DOIprefix}{doi:}
\providecommand{\ArXivprefix}{arXiv:}
\providecommand{\URLprefix}{URL: }
\providecommand{\Pubmedprefix}{pmid:}
\providecommand{\doi}[1]{\href{http://dx.doi.org/#1}{\path{#1}}}
\providecommand{\Pubmed}[1]{\href{pmid:#1}{\path{#1}}}
\providecommand{\bibinfo}[2]{#2}
\ifx\xfnm\relax \def\xfnm[#1]{\unskip,\space#1}\fi
\bibitem[{Atefi et~al.(2021)Atefi, Ge, Pitla and Schnable}]{Atefi2021-dw}
\bibinfo{author}{Atefi, A.}, \bibinfo{author}{Ge, Y.}, \bibinfo{author}{Pitla,
  S.}, \bibinfo{author}{Schnable, J.}, \bibinfo{year}{2021}.
\newblock \bibinfo{title}{Robotic technologies for high-throughput plant
  phenotyping: Contemporary reviews and future perspectives}.
\newblock \bibinfo{journal}{Front. Plant Sci.} \bibinfo{volume}{12},
  \bibinfo{pages}{611940}.
\bibitem[{Barron(2019)}]{barron2019generaladaptiverobustloss}
\bibinfo{author}{Barron, J.T.}, \bibinfo{year}{2019}.
\newblock \bibinfo{title}{A general and adaptive robust loss function}.
\newblock \URLprefix \url{https://arxiv.org/abs/1701.03077},
  \href{http://arxiv.org/abs/1701.03077}{{\tt arXiv:1701.03077}}.
\bibitem[{Calera et~al.(2023-06)Calera, Oliveira, Araujo, Filho, Toschi,
  Hernandes, Velasquez, Gasparino, Chowdhary, Higuti and
  Becker}]{ctx23878273330002401}
\bibinfo{author}{Calera, E.S.}, \bibinfo{author}{Oliveira, G.C.d.},
  \bibinfo{author}{Araujo, G.L.}, \bibinfo{author}{Filho, J.I.F.},
  \bibinfo{author}{Toschi, L.}, \bibinfo{author}{Hernandes, A.C.},
  \bibinfo{author}{Velasquez, A.E.B.}, \bibinfo{author}{Gasparino, M.V.},
  \bibinfo{author}{Chowdhary, G.}, \bibinfo{author}{Higuti, V.A.H.},
  \bibinfo{author}{Becker, M.}, \bibinfo{year}{2023-06}.
\newblock \bibinfo{title}{Under-canopy navigation for an agricultural rover
  based on image data}.
\newblock \bibinfo{journal}{Journal of intelligent \& robotic systems.}
  \bibinfo{volume}{108}.
\bibitem[{Carion et~al.(2020)Carion, Massa, Synnaeve, Usunier, Kirillov and
  Zagoruyko}]{detr}
\bibinfo{author}{Carion, N.}, \bibinfo{author}{Massa, F.},
  \bibinfo{author}{Synnaeve, G.}, \bibinfo{author}{Usunier, N.},
  \bibinfo{author}{Kirillov, A.}, \bibinfo{author}{Zagoruyko, S.},
  \bibinfo{year}{2020}.
\newblock \bibinfo{title}{End-to-end object detection with transformers}.
\newblock \URLprefix \url{https://arxiv.org/abs/2005.12872},
  \href{http://arxiv.org/abs/2005.12872}{{\tt arXiv:2005.12872}}.
\bibitem[{Cheppally et~al.(2023)Cheppally, Sharda and
  Wang}]{CHEPPALLY2023100182}
\bibinfo{author}{Cheppally, R.H.}, \bibinfo{author}{Sharda, A.},
  \bibinfo{author}{Wang, G.}, \bibinfo{year}{2023}.
\newblock \bibinfo{title}{Seed localization system suite with cnns for seed
  spacing estimation, population estimation and doubles identification}.
\newblock \bibinfo{journal}{Smart Agricultural Technology} \bibinfo{volume}{4},
  \bibinfo{pages}{100182}.
\newblock \URLprefix
  \url{https://www.sciencedirect.com/science/article/pii/S2772375523000126},
  \DOIprefix\doi{https://doi.org/10.1016/j.atech.2023.100182}.
\bibitem[{Contributors(2022)}]{mmengine2022}
\bibinfo{author}{Contributors, M.}, \bibinfo{year}{2022}.
\newblock \bibinfo{title}{{MMEngine}: Openmmlab foundational library for
  training deep learning models} .
\bibitem[{Dalal and Triggs(2005)}]{dalal2005histograms}
\bibinfo{author}{Dalal, N.}, \bibinfo{author}{Triggs, B.},
  \bibinfo{year}{2005}.
\newblock \bibinfo{title}{Histograms of oriented gradients for human
  detection}, in: \bibinfo{booktitle}{2005 IEEE computer society conference on
  computer vision and pattern recognition (CVPR'05)},
  \bibinfo{organization}{Ieee}. pp. \bibinfo{pages}{886--893}.
\bibitem[{Gan-Mor et~al.(2007)Gan-Mor, Clark and Upchurch}]{gan2007implement}
\bibinfo{author}{Gan-Mor, S.}, \bibinfo{author}{Clark, R.L.},
  \bibinfo{author}{Upchurch, B.L.}, \bibinfo{year}{2007}.
\newblock \bibinfo{title}{Implement lateral position accuracy under rtk-gps
  tractor guidance}.
\newblock \bibinfo{journal}{Computers and Electronics in Agriculture}
  \bibinfo{volume}{59}, \bibinfo{pages}{31--38}.
\bibitem[{Jiang et~al.(2016)Jiang, Wang, Wang and Liu}]{JIANG2016211}
\bibinfo{author}{Jiang, G.}, \bibinfo{author}{Wang, X.}, \bibinfo{author}{Wang,
  Z.}, \bibinfo{author}{Liu, H.}, \bibinfo{year}{2016}.
\newblock \bibinfo{title}{Wheat rows detection at the early growth stage based
  on hough transform and vanishing point}.
\newblock \bibinfo{journal}{Computers and Electronics in Agriculture}
  \bibinfo{volume}{123}, \bibinfo{pages}{211--223}.
\newblock \URLprefix
  \url{https://www.sciencedirect.com/science/article/pii/S0168169916300254},
  \DOIprefix\doi{https://doi.org/10.1016/j.compag.2016.02.002}.
\bibitem[{Kise et~al.(2002)Kise, Noguchi, Ishii and
  Terao}]{kise2002enhancement}
\bibinfo{author}{Kise, M.}, \bibinfo{author}{Noguchi, N.},
  \bibinfo{author}{Ishii, K.}, \bibinfo{author}{Terao, H.},
  \bibinfo{year}{2002}.
\newblock \bibinfo{title}{Enhancement of turning accuracy by path planning for
  robot tractor}, in: \bibinfo{booktitle}{Automation Technology for Off-Road
  Equipment Proceedings of the 2002 Conference},
  \bibinfo{organization}{American Society of Agricultural and Biological
  Engineers}. p. \bibinfo{pages}{398}.
\bibitem[{{Labelbox}(2024)}]{labelbox}
\bibinfo{author}{{Labelbox}}, \bibinfo{year}{2024}.
\newblock \bibinfo{title}{Labelbox}.
\newblock \bibinfo{howpublished}{\url{https://labelbox.com}}.
\newblock \bibinfo{note}{Online}.
\bibitem[{Li et~al.(2024)Li, Li, Wei and Wang}]{RowColRowDetection}
\bibinfo{author}{Li, B.}, \bibinfo{author}{Li, D.}, \bibinfo{author}{Wei, Z.},
  \bibinfo{author}{Wang, J.}, \bibinfo{year}{2024}.
\newblock \bibinfo{title}{Rethinking the crop row detection pipeline: An
  end-to-end method for crop row detection based on row-column attention}.
\newblock \bibinfo{journal}{Computers and Electronics in Agriculture}
  \bibinfo{volume}{225}, \bibinfo{pages}{109264}.
\newblock \URLprefix
  \url{https://www.sciencedirect.com/science/article/pii/S0168169924006550},
  \DOIprefix\doi{https://doi.org/10.1016/j.compag.2024.109264}.
\bibitem[{Li et~al.(2023)Li, Li, Kang, Feng, Long and Wang}]{LI2023120345}
\bibinfo{author}{Li, D.}, \bibinfo{author}{Li, B.}, \bibinfo{author}{Kang, S.},
  \bibinfo{author}{Feng, H.}, \bibinfo{author}{Long, S.},
  \bibinfo{author}{Wang, J.}, \bibinfo{year}{2023}.
\newblock \bibinfo{title}{E2cropdet: An efficient end-to-end solution to crop
  row detection}.
\newblock \bibinfo{journal}{Expert Systems with Applications}
  \bibinfo{volume}{227}, \bibinfo{pages}{120345}.
\newblock \URLprefix
  \url{https://www.sciencedirect.com/science/article/pii/S0957417423008473},
  \DOIprefix\doi{https://doi.org/10.1016/j.eswa.2023.120345}.
\bibitem[{Li et~al.(2021)Li, Huang, Sun and Liu}]{9624545}
\bibinfo{author}{Li, X.}, \bibinfo{author}{Huang, Z.}, \bibinfo{author}{Sun,
  X.}, \bibinfo{author}{Liu, T.}, \bibinfo{year}{2021}.
\newblock \bibinfo{title}{A fast detection method for polynomial fitting lane
  with self-attention module added}, in: \bibinfo{booktitle}{2021 International
  Conference on Control, Automation and Information Sciences (ICCAIS)}, pp.
  \bibinfo{pages}{46--51}.
\newblock \DOIprefix\doi{10.1109/ICCAIS52680.2021.9624545}.
\bibitem[{Liu et~al.(2022)Liu, Wang, Lu, Cao and
  Zhang}]{liu2022robustobjectdetectioninaccurate}
\bibinfo{author}{Liu, C.}, \bibinfo{author}{Wang, K.}, \bibinfo{author}{Lu,
  H.}, \bibinfo{author}{Cao, Z.}, \bibinfo{author}{Zhang, Z.},
  \bibinfo{year}{2022}.
\newblock \bibinfo{title}{Robust object detection with inaccurate bounding
  boxes}.
\newblock \URLprefix \url{https://arxiv.org/abs/2207.09697},
  \href{http://arxiv.org/abs/2207.09697}{{\tt arXiv:2207.09697}}.
\bibitem[{{Luxonis}({2024})}]{Luxonis}
\bibinfo{author}{{Luxonis}}, \bibinfo{year}{{2024}}.
\newblock \bibinfo{title}{Oak-d pro}.
\newblock \URLprefix \url{https://shop.luxonis.com/products/oak-d-pro}.
\bibitem[{Macenski et~al.(2022)Macenski, Foote, Gerkey, Lalancette and
  Woodall}]{ROS2}
\bibinfo{author}{Macenski, S.}, \bibinfo{author}{Foote, T.},
  \bibinfo{author}{Gerkey, B.}, \bibinfo{author}{Lalancette, C.},
  \bibinfo{author}{Woodall, W.}, \bibinfo{year}{2022}.
\newblock \bibinfo{title}{Robot operating system 2: Design, architecture, and
  uses in the wild}.
\newblock \bibinfo{journal}{Science Robotics} \bibinfo{volume}{7},
  \bibinfo{pages}{eabm6074}.
\newblock \URLprefix
  \url{https://www.science.org/doi/abs/10.1126/scirobotics.abm6074},
  \DOIprefix\doi{10.1126/scirobotics.abm6074}.
\bibitem[{Moraes~Rocha et~al.(2022)Moraes~Rocha, Vieira, Fonseca, Sousa,
  Pedrini and Soares}]{9885256}
\bibinfo{author}{Moraes~Rocha, B.}, \bibinfo{author}{Vieira, G.S.},
  \bibinfo{author}{Fonseca, A.U.}, \bibinfo{author}{Sousa, N.M.},
  \bibinfo{author}{Pedrini, H.}, \bibinfo{author}{Soares, F.},
  \bibinfo{year}{2022}.
\newblock \bibinfo{title}{Detection of curved rows and gaps in aerial images of
  sugarcane field using image processing techniques}.
\newblock \bibinfo{journal}{IEEE Canadian Journal of Electrical and Computer
  Engineering} \bibinfo{volume}{45}, \bibinfo{pages}{303--310}.
\newblock \DOIprefix\doi{10.1109/ICJECE.2022.3178749}.
\bibitem[{Mueller-Sim et~al.(2017)Mueller-Sim, Jenkins, Abel and
  Kantor}]{7989418}
\bibinfo{author}{Mueller-Sim, T.}, \bibinfo{author}{Jenkins, M.},
  \bibinfo{author}{Abel, J.}, \bibinfo{author}{Kantor, G.},
  \bibinfo{year}{2017}.
\newblock \bibinfo{title}{The robotanist: A ground-based agricultural robot for
  high-throughput crop phenotyping}, in: \bibinfo{booktitle}{2017 IEEE
  International Conference on Robotics and Automation (ICRA)}, pp.
  \bibinfo{pages}{3634--3639}.
\newblock \DOIprefix\doi{10.1109/ICRA.2017.7989418}.
\bibitem[{{NVIDIA Corporation, Santa Clara, CA, USA}(2025)}]{jetson}
\bibinfo{author}{{NVIDIA Corporation, Santa Clara, CA, USA}},
  \bibinfo{year}{2025}.
\newblock \bibinfo{title}{Nvidia jetson orin agx}.
\newblock \URLprefix \url{https://developer.nvidia.com/jetson}.
  \bibinfo{note}{accessed: 2025-08-13}.
\bibitem[{Panda et~al.(2023)Panda, Lee and Jawed}]{agronav}
\bibinfo{author}{Panda, S.K.}, \bibinfo{author}{Lee, Y.},
  \bibinfo{author}{Jawed, M.K.}, \bibinfo{year}{2023}.
\newblock \bibinfo{title}{Agronav: Autonomous navigation framework for
  agricultural robots and vehicles using semantic segmentation and semantic
  line detection}.
\newblock \URLprefix \url{https://arxiv.org/abs/2304.04333},
  \href{http://arxiv.org/abs/2304.04333}{{\tt arXiv:2304.04333}}.
\bibitem[{Rahmadian and Widyartono(2020)}]{9243253}
\bibinfo{author}{Rahmadian, R.}, \bibinfo{author}{Widyartono, M.},
  \bibinfo{year}{2020}.
\newblock \bibinfo{title}{Autonomous robotic in agriculture: A review}, in:
  \bibinfo{booktitle}{2020 Third International Conference on Vocational
  Education and Electrical Engineering (ICVEE)}, pp. \bibinfo{pages}{1--6}.
\newblock \DOIprefix\doi{10.1109/ICVEE50212.2020.9243253}.
\bibitem[{Rebuffi et~al.(2021)Rebuffi, Gowal, Calian, Stimberg, Wiles and
  Mann}]{rebuffi2021dataaugmentationimproverobustness}
\bibinfo{author}{Rebuffi, S.A.}, \bibinfo{author}{Gowal, S.},
  \bibinfo{author}{Calian, D.A.}, \bibinfo{author}{Stimberg, F.},
  \bibinfo{author}{Wiles, O.}, \bibinfo{author}{Mann, T.},
  \bibinfo{year}{2021}.
\newblock \bibinfo{title}{Data augmentation can improve robustness}.
\newblock \URLprefix \url{https://arxiv.org/abs/2111.05328},
  \href{http://arxiv.org/abs/2111.05328}{{\tt arXiv:2111.05328}}.
\bibitem[{Riba et~al.(2020)Riba, Mishkin, Ponsa, Rublee and
  Bradski}]{eriba2019kornia}
\bibinfo{author}{Riba, E.}, \bibinfo{author}{Mishkin, D.},
  \bibinfo{author}{Ponsa, D.}, \bibinfo{author}{Rublee, E.},
  \bibinfo{author}{Bradski, G.}, \bibinfo{year}{2020}.
\newblock \bibinfo{title}{Kornia: an open source differentiable computer vision
  library for pytorch}, in: \bibinfo{booktitle}{Winter Conference on
  Applications of Computer Vision}.
\newblock \URLprefix \url{https://arxiv.org/pdf/1910.02190.pdf}.
\bibitem[{Rocha et~al.(2020)Rocha, da~Silva~Vieira, Fonseca, Pedrini, de~Sousa
  and Soares}]{9255701}
\bibinfo{author}{Rocha, B.M.}, \bibinfo{author}{da~Silva~Vieira, G.},
  \bibinfo{author}{Fonseca, A.U.}, \bibinfo{author}{Pedrini, H.},
  \bibinfo{author}{de~Sousa, N.M.}, \bibinfo{author}{Soares, F.},
  \bibinfo{year}{2020}.
\newblock \bibinfo{title}{Evaluation and detection of gaps in curved sugarcane
  planting lines in aerial images}, in: \bibinfo{booktitle}{2020 IEEE Canadian
  Conference on Electrical and Computer Engineering (CCECE)}, pp.
  \bibinfo{pages}{1--4}.
\newblock \DOIprefix\doi{10.1109/CCECE47787.2020.9255701}.
\bibitem[{Sato and Chen(2022)}]{e2emetrics}
\bibinfo{author}{Sato, T.}, \bibinfo{author}{Chen, Q.A.}, \bibinfo{year}{2022}.
\newblock \bibinfo{title}{Towards driving-oriented metric for lane detection
  models}.
\newblock \URLprefix \url{https://arxiv.org/abs/2203.16851},
  \href{http://arxiv.org/abs/2203.16851}{{\tt arXiv:2203.16851}}.
\bibitem[{Satzoda and Trivedi(2014)}]{LPD}
\bibinfo{author}{Satzoda, R.K.}, \bibinfo{author}{Trivedi, M.M.},
  \bibinfo{year}{2014}.
\newblock \bibinfo{title}{On performance evaluation metrics for lane
  estimation}, in: \bibinfo{booktitle}{2014 22nd International Conference on
  Pattern Recognition}, pp. \bibinfo{pages}{2625--2630}.
\newblock \DOIprefix\doi{10.1109/ICPR.2014.453}.
\bibitem[{de~Silva et~al.(2024)de~Silva, Cielniak, Wang and Gao}]{CRDLD}
\bibinfo{author}{de~Silva, R.}, \bibinfo{author}{Cielniak, G.},
  \bibinfo{author}{Wang, G.}, \bibinfo{author}{Gao, J.}, \bibinfo{year}{2024}.
\newblock \bibinfo{title}{Deep learning-based crop row detection for infield
  navigation of agri-robots}.
\newblock \bibinfo{journal}{Journal of Field Robotics} \bibinfo{volume}{41},
  \bibinfo{pages}{2299--2321}.
\newblock \URLprefix
  \url{https://onlinelibrary.wiley.com/doi/abs/10.1002/rob.22238},
  \DOIprefix\doi{10.1002/rob.22238},
  \href{http://arxiv.org/abs/https://onlinelibrary.wiley.com/doi/pdf/10.1002/rob.22238}{{\tt
  arXiv:https://onlinelibrary.wiley.com/doi/pdf/10.1002/rob.22238}}.
\bibitem[{Sivakumar et~al.(2021)Sivakumar, Modi, Gasparino, Ellis,
  Baquero~Velasquez, Chowdhary and Gupta}]{Sivakumar_2021}
\bibinfo{author}{Sivakumar, A.}, \bibinfo{author}{Modi, S.},
  \bibinfo{author}{Gasparino, M.}, \bibinfo{author}{Ellis, C.},
  \bibinfo{author}{Baquero~Velasquez, A.}, \bibinfo{author}{Chowdhary, G.},
  \bibinfo{author}{Gupta, S.}, \bibinfo{year}{2021}.
\newblock \bibinfo{title}{Learned visual navigation for under-canopy
  agricultural robots}, in: \bibinfo{booktitle}{Robotics: Science and Systems
  XVII}, \bibinfo{publisher}{Robotics: Science and Systems Foundation}.
\newblock \URLprefix \url{http://dx.doi.org/10.15607/RSS.2021.XVII.019},
  \DOIprefix\doi{10.15607/rss.2021.xvii.019}.
\bibitem[{Sivakumar et~al.(2024)Sivakumar, Gasparino, McGuire, Higuti, Akcal
  and Chowdhary}]{sivakumar2024lessons}
\bibinfo{author}{Sivakumar, A.N.}, \bibinfo{author}{Gasparino, M.V.},
  \bibinfo{author}{McGuire, M.}, \bibinfo{author}{Higuti, V.A.H.},
  \bibinfo{author}{Akcal, M.U.}, \bibinfo{author}{Chowdhary, G.},
  \bibinfo{year}{2024}.
\newblock \bibinfo{title}{Lessons from deploying cropfollow++: Under-canopy
  agricultural navigation with keypoints}.
\newblock \href{http://arxiv.org/abs/2404.17718}{{\tt arXiv:2404.17718}}.
\bibitem[{Tabelini et~al.(2020)Tabelini, Berriel, Paixão, Badue, Souza and
  Oliveira-Santos}]{tabelini2020polylanenet}
\bibinfo{author}{Tabelini, L.}, \bibinfo{author}{Berriel, R.},
  \bibinfo{author}{Paixão, T.M.}, \bibinfo{author}{Badue, C.},
  \bibinfo{author}{Souza, A.F.D.}, \bibinfo{author}{Oliveira-Santos, T.},
  \bibinfo{year}{2020}.
\newblock \bibinfo{title}{Polylanenet: Lane estimation via deep polynomial
  regression}.
\newblock \href{http://arxiv.org/abs/2004.10924}{{\tt arXiv:2004.10924}}.
\bibitem[{Tan and Le(2019)}]{tan2019efficientnet}
\bibinfo{author}{Tan, M.}, \bibinfo{author}{Le, Q.}, \bibinfo{year}{2019}.
\newblock \bibinfo{title}{Efficientnet: Rethinking model scaling for
  convolutional neural networks}, in: \bibinfo{booktitle}{International
  conference on machine learning}, \bibinfo{organization}{PMLR}. pp.
  \bibinfo{pages}{6105--6114}.
\bibitem[{Targ et~al.(2016)Targ, Almeida and Lyman}]{targ2016resnet}
\bibinfo{author}{Targ, S.}, \bibinfo{author}{Almeida, D.},
  \bibinfo{author}{Lyman, K.}, \bibinfo{year}{2016}.
\newblock \bibinfo{title}{Resnet in resnet: Generalizing residual
  architectures}.
\newblock \bibinfo{journal}{arXiv preprint arXiv:1603.08029} .
\bibitem[{Tu et~al.(2014)Tu, van Wyk, Djouani, Hamam and Du}]{7003860}
\bibinfo{author}{Tu, C.}, \bibinfo{author}{van Wyk, B.J.},
  \bibinfo{author}{Djouani, K.}, \bibinfo{author}{Hamam, Y.},
  \bibinfo{author}{Du, S.}, \bibinfo{year}{2014}.
\newblock \bibinfo{title}{An efficient crop row detection method for
  agriculture robots}, in: \bibinfo{booktitle}{2014 7th International Congress
  on Image and Signal Processing}, pp. \bibinfo{pages}{655--659}.
\newblock \DOIprefix\doi{10.1109/CISP.2014.7003860}.
\bibitem[{TuSimple(2017)}]{tusimple}
\bibinfo{author}{TuSimple}, \bibinfo{year}{2017}.
\newblock \bibinfo{title}{Tusimple lane detection challenge}.
\newblock
  \bibinfo{howpublished}{\url{https://github.com/TuSimple/tusimple-benchmark/tree/master/doc/lane\_detection}}.
\newblock \bibinfo{note}{Accessed: 2024-11-14}.
\bibitem[{Xu et~al.(2022)Xu, Pan, Pan, Hoi, Yi and Xu}]{xu2022regnet}
\bibinfo{author}{Xu, J.}, \bibinfo{author}{Pan, Y.}, \bibinfo{author}{Pan, X.},
  \bibinfo{author}{Hoi, S.}, \bibinfo{author}{Yi, Z.}, \bibinfo{author}{Xu,
  Z.}, \bibinfo{year}{2022}.
\newblock \bibinfo{title}{Regnet: self-regulated network for image
  classification}.
\newblock \bibinfo{journal}{IEEE Transactions on Neural Networks and Learning
  Systems} .
\bibitem[{Yan et~al.(2021)Yan, Wu, Xiao, Mei and Meng}]{yan_seed_mapping}
\bibinfo{author}{Yan, B.}, \bibinfo{author}{Wu, G.}, \bibinfo{author}{Xiao,
  Y.}, \bibinfo{author}{Mei, H.}, \bibinfo{author}{Meng, Z.},
  \bibinfo{year}{2021}.
\newblock \bibinfo{title}{Development and evaluation of a seed position mapping
  system}.
\newblock \bibinfo{journal}{Computers and Electronics in Agriculture}
  \bibinfo{volume}{190}, \bibinfo{pages}{106446}.
\newblock \URLprefix
  \url{https://www.sciencedirect.com/science/article/pii/S0168169921004634},
  \DOIprefix\doi{https://doi.org/10.1016/j.compag.2021.106446}.
\bibitem[{Yang et~al.(2024)Yang, Tian, You, Jia, Liu, Pan and
  John}]{yang2024polylanenet++}
\bibinfo{author}{Yang, C.}, \bibinfo{author}{Tian, Z.}, \bibinfo{author}{You,
  X.}, \bibinfo{author}{Jia, K.}, \bibinfo{author}{Liu, T.},
  \bibinfo{author}{Pan, Z.}, \bibinfo{author}{John, V.}, \bibinfo{year}{2024}.
\newblock \bibinfo{title}{Polylanenet++: enhancing the polynomial regression
  lane detection based on spatio-temporal fusion}.
\newblock \bibinfo{journal}{Signal, Image and Video Processing} ,
  \bibinfo{pages}{1--10}.
\bibitem[{Yang et~al.(2023)Yang, Zhou, Yue, Zhang, Wen, Ma, Xu and
  Chen}]{yang2023real}
\bibinfo{author}{Yang, Y.}, \bibinfo{author}{Zhou, Y.}, \bibinfo{author}{Yue,
  X.}, \bibinfo{author}{Zhang, G.}, \bibinfo{author}{Wen, X.},
  \bibinfo{author}{Ma, B.}, \bibinfo{author}{Xu, L.}, \bibinfo{author}{Chen,
  L.}, \bibinfo{year}{2023}.
\newblock \bibinfo{title}{Real-time detection of crop rows in maize fields
  based on autonomous extraction of roi}.
\newblock \bibinfo{journal}{Expert Systems with Applications}
  \bibinfo{volume}{213}, \bibinfo{pages}{118826}.
\bibitem[{Yang et~al.(2022)Yang, Yang, Li, Zhou, Zhang, Yu and Liu}]{tasseled}
\bibinfo{author}{Yang, Z.}, \bibinfo{author}{Yang, Y.}, \bibinfo{author}{Li,
  C.}, \bibinfo{author}{Zhou, Y.}, \bibinfo{author}{Zhang, X.},
  \bibinfo{author}{Yu, Y.}, \bibinfo{author}{Liu, D.}, \bibinfo{year}{2022}.
\newblock \bibinfo{title}{Tasseled crop rows detection based on micro-region of
  interest and logarithmic transformation}.
\newblock \bibinfo{journal}{Frontiers in Plant Science} \bibinfo{volume}{Volume
  13 - 2022}.
\newblock \URLprefix
  \url{https://www.frontiersin.org/journals/plant-science/articles/10.3389/fpls.2022.916474},
  \DOIprefix\doi{10.3389/fpls.2022.916474}.
\bibitem[{Yu et~al.(2023)Yu, Zhang, Shu, Chen, Chen, Yang, Tang and
  Zhang}]{YU2023107811}
\bibinfo{author}{Yu, J.}, \bibinfo{author}{Zhang, J.}, \bibinfo{author}{Shu,
  A.}, \bibinfo{author}{Chen, Y.}, \bibinfo{author}{Chen, J.},
  \bibinfo{author}{Yang, Y.}, \bibinfo{author}{Tang, W.},
  \bibinfo{author}{Zhang, Y.}, \bibinfo{year}{2023}.
\newblock \bibinfo{title}{Study of convolutional neural network-based semantic
  segmentation methods on edge intelligence devices for field agricultural
  robot navigation line extraction}.
\newblock \bibinfo{journal}{Computers and Electronics in Agriculture}
  \bibinfo{volume}{209}, \bibinfo{pages}{107811}.
\newblock \URLprefix
  \url{https://www.sciencedirect.com/science/article/pii/S0168169923001990},
  \DOIprefix\doi{https://doi.org/10.1016/j.compag.2023.107811}.
\bibitem[{Zhao et~al.(2024a)Zhao, Yuan, Yang and Zhang}]{Yuan}
\bibinfo{author}{Zhao, R.}, \bibinfo{author}{Yuan, X.}, \bibinfo{author}{Yang,
  Z.}, \bibinfo{author}{Zhang, L.}, \bibinfo{year}{2024}a.
\newblock \bibinfo{title}{Image-based crop row detection utilizing the hough
  transform and dbscan clustering analysis}.
\newblock \bibinfo{journal}{IET Image Processing} \bibinfo{volume}{18},
  \bibinfo{pages}{1161--1177}.
\newblock \URLprefix
  \url{https://ietresearch.onlinelibrary.wiley.com/doi/abs/10.1049/ipr2.13016},
  \DOIprefix\doi{https://doi.org/10.1049/ipr2.13016},
  \href{http://arxiv.org/abs/https://ietresearch.onlinelibrary.wiley.com/doi/pdf/10.1049/ipr2.13016}{{\tt
  arXiv:https://ietresearch.onlinelibrary.wiley.com/doi/pdf/10.1049/ipr2.13016}}.
\bibitem[{Zhao et~al.(2024b)Zhao, Lv, Xu, Wei, Wang, Dang, Liu and
  Chen}]{rtdetr}
\bibinfo{author}{Zhao, Y.}, \bibinfo{author}{Lv, W.}, \bibinfo{author}{Xu, S.},
  \bibinfo{author}{Wei, J.}, \bibinfo{author}{Wang, G.}, \bibinfo{author}{Dang,
  Q.}, \bibinfo{author}{Liu, Y.}, \bibinfo{author}{Chen, J.},
  \bibinfo{year}{2024}b.
\newblock \bibinfo{title}{Detrs beat yolos on real-time object detection}.
\newblock \URLprefix \url{https://arxiv.org/abs/2304.08069},
  \href{http://arxiv.org/abs/2304.08069}{{\tt arXiv:2304.08069}}.
\bibitem[{Zhu et~al.(2021)Zhu, Su, Lu, Li, Wang and
  Dai}]{zhu2021deformabledetrdeformabletransformers}
\bibinfo{author}{Zhu, X.}, \bibinfo{author}{Su, W.}, \bibinfo{author}{Lu, L.},
  \bibinfo{author}{Li, B.}, \bibinfo{author}{Wang, X.}, \bibinfo{author}{Dai,
  J.}, \bibinfo{year}{2021}.
\newblock \bibinfo{title}{Deformable detr: Deformable transformers for
  end-to-end object detection}.
\newblock \URLprefix \url{https://arxiv.org/abs/2010.04159},
  \href{http://arxiv.org/abs/2010.04159}{{\tt arXiv:2010.04159}}.

\end{thebibliography}
\appendix
\section{Backbone Performance on Various Testsets}\label{sec:backbone_performance_appendix}
The performance of the RowDetr model on various testsets is shown in Table~\ref{tab:sourghum_results}, Table~\ref{tab:crdl_results}, Table~\ref{tab:v5_residue_results}, and Table~\ref{tab:v4_v6_weeds_results}.
\begin{table}[ht]
    \centering
    \caption{Results on Sourghum Test set}
    \begin{tabular}{lrrrr}
        \hline
         Backbone & LPD & TuSimple FNR & TuSimple FPR & TuSimple F1 \\
         \hline
        RowDetr[efficientnet] & \textbf{0.91} & 0.022 & \textbf{0.755} & \textbf{0.385} \\
        RowDetr[resnet18] & 0.93  & 0.0204 & 0.763 & 0.376 \\
        RowDetr[regnetx\_008] & 0.93  & \textbf{0.017} & 0.774 & 0.362 \\
        RowDetr[resnet50] & 0.93 & 0.019 & 0.757 & 0.384 \\
\hline
    \end{tabular}
    \label{tab:sourghum_results}          
\end{table}

\begin{table}[ht]
    \centering
    \caption{Results on CRDL Test set}
    \begin{tabular}{lrrrr}
        \hline
         Model & LPD & TuSimple FNR & TuSimple FPR & TuSimple F1 \\
         \hline
         
         RowDetr[efficientnet] & 0.189509 & \textbf{0.052} & 0.217 & 0.853 \\
         RowDetr[resnet18] & 0.184209  & 0.060 & \textbf{0.199} & \textbf{0.860} \\
         RowDetr[regnetx\_008] & \textbf{0.173872}  & 0.059 & 0.211 & 0.853 \\
         RowDetr[resnet50]  & 0.205441 & 0.063 & 0.211 & 0.851 \\
         \hline
         \end{tabular}
        \label{tab:crdl_results}          
\end{table}

\begin{table}[ht]
    \centering
    \caption{Results on corn V5 with Residue}
    \begin{tabular}{lrrrr}
        \hline
         Model & LPD & TuSimple FNR & TuSimple FPR & TuSimple F1 \\
         \hline
         
         RowDetr[efficientnet] & 0.229 & 0.0 & 0.529 & 0.640 \\
         RowDetr[resnet18] & 0.237  & 0.0 & 0.0 & 0.591 \\
         RowDetr[regnetx\_008] & \textbf{0.226}  & 0.0 & \textbf{0.497} & \textbf{0.668} \\
         RowDetr[resnet50]  & 0.230 & 0.06 & 0.582 & 0.588 \\
         \hline
         \end{tabular}
        \label{tab:v5_residue_results}          
\end{table}

\begin{table}[ht]
    \centering
    \caption{Results on corn V4 \& corn V6 with weeds}
    \begin{tabular}{lrrrr}
        \hline
         Model & LPD & TuSimple FNR & TuSimple FPR & TuSimple F1 \\
         \hline
         
         RowDetr[efficientnet] & 0.234 & 0.038 & 0.496 & 0.652 \\
         RowDetr[resnet18] & 0.240  & 0.039 & \textbf{0.494} & \textbf{0.653} \\
         RowDetr[regnetx\_008] & \textbf{0.229}  & 0.038 & 0.516 & 0.635 \\
         RowDetr[resnet50]  & 0.252 & 0.038 & 0.568 & 0.587 \\
         \hline
         \end{tabular}
        \label{tab:v4_v6_weeds_results}          
\end{table} 

\end{document}